%% file: Unified-Measure.tex
\documentclass[twoside,11pt]{article}

\usepackage{blindtext}
\usepackage{amsmath}
\usepackage{tikz}
\usetikzlibrary{arrows.meta,positioning,fit,backgrounds}

\usepackage[preprint]{jmlr2e}

\usepackage{lastpage}

\jmlrheading{Preprint}{2026}{1-59}{Draft}{May 2026}{}{Ranganath and Singhal}

\firstpageno{1}

\title{A Unified Measure-Theoretic View of Diffusion, Score-Based, and Flow Matching Generative Models}

\author{\name Aditya Ranganath \email ranganath2@llnl.gov \\
       \addr Center for Applied Scientific Computing\\
       Lawrence Livermore National Laboratory\\
       Livermore, CA 94551, USA
       \AND
       \name Mukesh Singhal \email msinghal@ucmerced.edu \\
       \addr Department of Electrical Engineering\\
       University of California\\
       Merced, CA 95343, USA}
\editor{}

\begin{document}

\maketitle
\makeatother

\begin{abstract}
We survey continuous-time generative modeling methods based on transporting a simple reference distribution to a data distribution via stochastic or deterministic dynamics. We present a unified framework in which diffusion models, score-based generative models, and flow matching are instances of learning a time-dependent vector field that induces a family of marginals 
$(\rho_t)_{t\in[0,1]}$ governed by a continuity/Fokker–Planck equation. Within this framework, we (i) derive reverse-time sampling for diffusion/score models as controlled stochastic dynamics, (ii) show the probability flow ODE yields identical marginals and connects diffusion to likelihood-based normalizing flows, and (iii) interpret flow matching as direct regression of the velocity field under a chosen interpolation, clarifying when it coincides with (or differs from) score-based training. We compare objectives, sampling schemes, and discretization errors under unified notation, discuss connections to Schrödinger bridges and entropic optimal transport, and summarize theoretical guarantees and open problems on approximation, stability, and scalability.
\end{abstract}


\section{Introduction and Reading Guide}
\input{sections/section1_introduction}
\label{sec:intro}

\section{Unified Notation and Preliminaries}
\input{sections/section2_preliminaries}
\label{sec:prelim}

\section{Forward Processes and Probability Paths}
\label{sec:forward}
\input{sections/section3_forward_process}

\section{Reverse-Time Dynamics and Sampling}
\label{sec:reverse}
\input{sections/section4_reverse_dynamics}

\section{Training Objectives and Equivalences}
\label{sec:objectives}
\input{sections/section5_objectives}

\section{Probability Flow ODE and Likelihood Connections}
\label{sec:pfode}
\input{sections/section6_probability_flow_ode}

\section{Flow Matching and Rectified Flows}
\label{sec:flowmatching}
\input{sections/section7_flow_matching}

\section{Unified Comparison}
\label{sec:comparison}
\input{sections/section8_unified}

\section{Theory and Error Decomposition}
\label{sec:theory}
\input{sections/section9_theory}

\section{Conclusion: Open Problems and Research Directions}
\label{sec:open}
\input{sections/section10_open_problems}

\newpage

\appendix

\section{Reverse-Time Diffusions and the Score Term}

\input{sections/appendixA_reverse_time_diffusions}
\label{app:reverse_time}

\section{Fokker--Planck, Continuity Equations, and the Probability-Flow ODE}
\input{sections/appendixB_pfode_and_pdes}
\label{app:pfode_pdes}

\section{Objective Equivalences: Denoising Score Matching, DDPM Losses, and Weighted Fisher Divergence}
\input{sections/appendixC_objective_equivalences}
\label{app:objective_equivalences}

\section{Flow Matching: Conditional Targets and Marginal Optimal Velocity}\label{app:flow_matching}
\input{sections/appendixD_flow_matching_derivations}

\section{Measure-Theoretic Background: Probability Laws, Pushforwards, and Path Measures}\label{app:measure_background}
\input{sections/appendixE_measure_theoretic_background}

\section{Schr\"odinger Bridges, Entropic Transport, and Connections to Diffusion Models}\input{sections/appendixF_schrodinger_bridges}\label{app:sb}



\bibliography{refs}

\end{document}

%% file: sections/section1_introduction.tex
Generative modeling seeks to learn mechanisms for producing samples from a complex data distribution $\rho_{\mathrm{data}}$ on $\mathbb{R}^d$. Beyond unconditional synthesis, modern generative models serve as priors and proposal mechanisms in downstream tasks, including conditional generation and inverse problems. Over the past decade, the field has progressed through several paradigms---variational autoencoders, generative adversarial networks, and likelihood-based normalizing flows \citep{kingma2014vae,rezende2014vae,goodfellow2014gan,dinh2015nice,dinh2017realnvp,papamakarios2021normalizing}---toward a family of methods that construct samples by evolving a simple reference distribution through time-dependent dynamics. Diffusion models and closely related score-based generative models have become influential approaches to high-fidelity synthesis in high dimensions, with diffusion probabilistic modeling popularized in modern deep learning by denoising diffusion probabilistic models (DDPM) \citep{ho2020ddpm} and its successors \citep{song2020ddim,nichol2021improved,kingma2021vdm,karras2022edm}. More recent work has broadened this design space further through deterministic degradations, stochastic interpolants, consistency-style models, Bayesian flow networks, and discrete diffusion alternatives \citep{bansal2022cold,albergo2025stochastic,song2023consistency,kim2024ctm,graves2023bfn,lou2024sedd}.

\paragraph{What does it mean for a model to be \emph{generative}?}
A model is called \emph{generative} if it specifies, explicitly or implicitly, a mechanism for producing new samples that resemble draws from an unknown data distribution $\rho_{\mathrm{data}}$. Formally, given data $x\in\mathbb{R}^d$ (or more general sample spaces), the goal is to learn a distribution $p_\theta(x)$ such that $p_\theta \approx \rho_{\mathrm{data}}$ in a meaningful sense. However, being generative does not require an explicit closed-form density. What matters is that the model defines a sampling procedure that maps randomness to data, for example
\[
x = G_\theta(z), \qquad z \sim p(z),
\]
where $p(z)$ is a simple base distribution, such as a standard Gaussian, and $G_\theta$ is a learned transformation. This view emphasizes that generative modeling can be interpreted as a problem of learning \emph{probability transport}: pushing a simple reference distribution through learned dynamics to match the data distribution \citep{goodfellow2016deep,papamakarios2021normalizing}.

\paragraph{Background: a brief map of generative modeling paradigms.}
Before diffusion and score-based methods became widely adopted, three families of deep generative models shaped the modern literature. These paradigms differ mainly in how they represent and learn the model distribution $p_\theta$.

\emph{Latent-variable likelihood models} such as variational autoencoders (VAEs) learn an explicit generative model together with an amortized inference network by maximizing a variational lower bound on the data log-likelihood \citep{kingma2014vae,rezende2014vae}. VAEs offer stable training and likelihood-based evaluation, but classic formulations can trade off sample sharpness against generalizability, resulting in richer decoders and more narrow, yet, expressive priors.

\emph{Implicit generative models} such as generative adversarial networks (GANs) learn to generate samples through a minimax objective against a discriminator \citep{goodfellow2014gan}. GANs produce high-fidelity samples and somewhat avoid explicit likelihood computation when compared with latent-variable likelihood models, but training can be unstable and evaluation can be challenging \citep{arjovsky2017towards,mescheder2018converge,salimans2016improved}. Mode dropping and sensitivity to hyperparameters are recurring issues.

\emph{Normalizing flows} learn an invertible map that transports a simple base distribution (e.g., Gaussian) to the data distribution, enabling exact likelihood computation via change-of-variables \citep{dinh2015nice,dinh2017realnvp,papamakarios2021normalizing}. Flows provide a clear connection to transport and Jacobian-based likelihoods, but architectural constraints required for tractable Jacobians can limit expressivity or increase compute in high dimensions.

Taken together, these paradigms highlight recurring design choices: whether the model has an explicit likelihood, whether generation is defined through an invertible map or an implicit sampler, and how training balances sample quality, coverage, and tractability \citep{goodfellow2016deep,papamakarios2021normalizing}. Diffusion and score-based models can be viewed as inheriting aspects of all three traditions: like flows, they admit a transport interpretation; like implicit models, they emphasize flexible samplers and high sample quality; and like latent-variable likelihood models, they often come with likelihood-based objectives or variational bounds \citep{ho2020ddpm,song2021sde,kingma2021vdm}. On the contrary, more recent literature has explored substantially different ways of instantiating the same generative goal, including deterministic degradations in place of Gaussian noising, stochastic interpolants, consistency-based one-step or few-step generation, Bayesian flow-network formulations, and ratio-based discrete diffusion methods \citep{bansal2022cold,albergo2025stochastic,song2023consistency,kim2024ctm,graves2023bfn,lou2024sedd}.

A second example---especially relevant to scientific and engineering applications---is solving inverse problems such as deblurring, inpainting, MRI/CT reconstruction, or more general linear and nonlinear measurement models of the form $y=\mathcal{A}(x)+\varepsilon$. Here, generative models generate priors for recovering $x$ from partial or noisy observations $y$. In diffusion and score-based formulations, the learned score field provides a principled way to construct posterior sampling by combining a prior score with a likelihood term, leading to algorithms for conditional generation and reconstruction. This can be interpreted as predictor--corrector approach or approximate posterior samplers \citep{song2021sde,jalal2021robust,chung2022score,chung2023dps,kawar2022ddrm,tewari2023forwardmodels,rout2023latentinverse,li2024diffoptcontrol,janati2024dcps}. Closely related ideas have also been used for guided image editing and inpainting, where the conditioning signal acts as a soft constraint along the transport path rather than as an explicit post-transport correction \citep{meng2022sdedit,lugmayr2022repaint,corneanu2024latentpaint,zhang2023ucdir}. In this setting, the choice of sampler (reverse-time SDE versus probability-flow ODE), the time discretization, and the weighting induced by the training objective can strongly affect reconstruction fidelity and stability. This bolsters our motivation for a unified theory of objectives and samplers. The same learned model can behave very differently in inverse problems depending on how its dynamics are discretized and how conditioning is imposed.

\paragraph{Historical narrative.}
Modern diffusion modeling in deep learning was popularized by DDPM, which framed generation as reversing a discrete-time Markov noising process trained by a variational/denoising objective \citep{sohldickstein2015deep,ho2020ddpm}. A continuous-time reformulation then unified several diffusion-like constructions by expressing the forward noising process as an SDE and deriving a reverse-time SDE whose drift depends on the time-dependent score. This SDE perspective also made an explicit deterministic probability flow ODE, sharing the same marginals as the stochastic diffusion and connect diffusion sampling to continuous normalizing flows\citep{song2021sde,chen2018neuralode,grathwohl2019ffjord}. More recently, flow matching reframed training as simulation-free regression of vector fields for prescribed probability paths, including diffusion paths as special cases, enabling scalable training of continuous-normalizing-flow-style generators with standard ODE solvers \citep{lipman2022flowmatching}. In parallel, straightening variants such as rectified flow learnt transport dynamics whose trajectories are as close to straight lines as possible, yielding accurate generation with very coarse discretization\citep{liu2022rectified}. Other work proposed alternative routes to transport-based generation, including autoregressive diffusion, non-Gaussian degradation processes, and path-space constructions tailored to discrete or constrained settings \citep{hoogeboom2022ardm,bansal2022cold,campbell2022ctddm}.

\input{figures/fig_unified_transport}

Figure~\ref{fig:unified_transport} summarizes the common transport perspective that motivates this survey. It shows how diffusion, score-based sampling, probability-flow ODEs, and flow matching can be organized around three design choices: the probability path, the learned field, and the sampling dynamics.

\subsection{Unifying theme: probability transport through learned fields}\label{subsec:theme}

This survey develops a unified technical view: diffusion models, score-based generative models, and flow matching can be interpreted as instances of learning probability transport \citep{song2021sde,lipman2022flowmatching}. We enumerate the integral elements of this survey below:
\begin{enumerate}
\item A family of intermediate distributions $(\rho_t)_{t\in[0,1]}$ connecting a complex target $\rho_0$ (data) to a tractable reference $\rho_1$ (often Gaussian).
\item A time-dependent field that determines how probability mass evolves along this path:
\begin{itemize}
\item a \emph{score field} $s_t(x)=\nabla_x\log\rho_t(x)$, used prominently in score-based and diffusion formulations, or
\item a \emph{velocity field} $v_t(x)$ defining deterministic transport via a continuity equation, used prominently in CNF and flow-matching formulations.
\end{itemize}
\end{enumerate}

In score-based generative modeling, the reverse-time sampler obtains time-dependent scores and integrates a reverse-time SDE; in addition, an associated deterministic probability flow ODE can be constructed that shares the same marginals as the SDE and enables likelihood computation via change-of-variables along the ODE flow \citep{song2021sde,chen2018neuralode,grathwohl2019ffjord}. The roots of this perspective traces back to score matching---a method for learning unnormalized models by matching score functions---together with denoising score matching, which connects score estimation to denoising objectives \citep{hyvarinen2005score,vincent2011dsm,song2020ssm,song2020improved}. These connections are explicitly leveraged in modern diffusion-model training objectives, including the DDPM formulation \citep{ho2020ddpm,kingma2021vdm}.

\subsection{Why a unifying survey is needed now}\label{subsec:why}

A unifying survey is timely for at least three reasons.

\paragraph{(i) Methodological convergence.}
The boundary between diffusion/score and flow-based methods have blurred. Score-based modeling through SDEs explicitly derives both a reverse-time SDE sampler and an equivalent deterministic ODE sampler (probability flow ODE), thereby connecting diffusion-style modeling with continuous normalizing flows and ODE-based likelihood computation; flow matching makes this connection even more explicit by learning velocity fields directly along prescribed probability paths \citep{song2021sde,chen2018neuralode,grathwohl2019ffjord,lipman2022flowmatching}. More recent conditional and inverse-problem formulations further suggest that these methods are converging operationally even when they differ in parameterization \citep{tewari2023forwardmodels,li2024diffoptcontrol}.

\paragraph{(ii) Fragmented notation and competing derivations.}
Diffusion models are often introduced through discrete-time Markov chains and variational training objectives, while score-based approaches are presented through SDE time reversal, and flow matching is presented through regression of vector fields over probability paths. These presentations can obscure the fact that many techniques differ mainly by (a) the choice of path $\rho_t$, (b) whether one learns scores or velocities, and (c) whether one samples by SDE or ODE integration \citep{ho2020ddpm,song2021sde,lipman2022flowmatching,kingma2021vdm}.

\paragraph{(iii) Practical stakes: sampling, stability, and compute.}
Once methods are viewed as learned dynamics, numerical analysis and modeling choices become critical: discretization bias, stiffness, solver choice, and the role of stochasticity (SDE versus ODE) materially affect sample quality, cost, and robustness. The neural ODE/CNF literature provides a framework for understanding likelihood computation and the numerical behavior of continuous-time flows, which becomes directly relevant when diffusion models are sampled with deterministic ODEs \citep{chen2018neuralode,grathwohl2019ffjord,song2021sde,karras2022edm}. The same issues become even more visible in inverse-problem settings, where conditioning constraints can magnify small numerical or modeling errors \citep{rout2023latentinverse,pandey2024fastinverse}.

\subsection{Scope and goals}\label{subsec:scope}

\paragraph{Covered.}
We focus on continuous-time formulations of:
\begin{itemize}
\item diffusion probabilistic models and score-based generative models (forward SDEs, reverse-time SDE sampling, denoising score matching objectives, and probability flow ODE sampling) \citep{ho2020ddpm,song2021sde,kingma2021vdm};
\item flow matching as a method for training CNFs by regressing vector fields of fixed probability paths, including diffusion paths as special instances \citep{lipman2022flowmatching}; and
\item connections to Schr\"odinger bridges / entropic optimal transport on path space as an interpretation and generalization of score-based diffusion methods \citep{debortoli2021dsb,leonard2014survey}.
\end{itemize}

\paragraph{Not covered in depth.}
We do not attempt a comprehensive catalog of architectures or application domains; our emphasis is on mathematical unification, objective equivalences, and sampling dynamics, together with the consequences of design choices such as paths and parameterizations.

\subsection{Related surveys and tutorials}\label{subsec:related_surveys}

Several recent survey and tutorial papers partially overlap with our scope, but each emphasizes a different slice of the landscape. Broad diffusion surveys by Yang et al., Cao et al., and Ahsan et al. focus primarily on diffusion-model foundations, algorithmic variants, and applications across modalities and domains \citep{yang2022diffsurvey,cao2022survey,ahsan2024comprehensive}. Efficiency-oriented reviews by Ma et al. and Shen et al. concentrate on acceleration, efficient training and inference, and deployment considerations for diffusion models \citep{ma2024efficient,shen2025efficient}. Tang and Zhao provide a technical tutorial centered specifically on score-based diffusion models through the stochastic-differential-equation formulation \citep{tang2024sdetutorial}, while Lipman et al. give a comprehensive review of flow matching and its extensions \citep{lipman2024fmguide}. Holderrieth and Erives offer a first-principles tutorial spanning both diffusion and flow matching, with emphasis on practical construction of modern generators \citep{holderrieth2025intro}.

Our survey differs in emphasis and organization. Instead of focusing only diffusion models, only flow matching, or only efficiency, we adopt a unified measure-theoretic probability-transport umbrella in which diffusion, score-based models, probability-flow ODEs, and flow matching are treated within a common path--field--sampler framework. We also emphasize formal equivalence statements, comparative structure, and supporting appendices on reverse-time diffusions, objective equivalences, path measures, and Schr\"odinger bridges. In that sense, the present paper is intended to complement existing diffusion-heavy surveys and method-specific tutorials by offering a more explicitly unifying and mathematically organized perspective \citep{yang2022diffsurvey,cao2022survey,ahsan2024comprehensive,tang2024sdetutorial,lipman2024fmguide,holderrieth2025intro}.

\subsection{A map of design choices}\label{subsec:map}

We organize the landscape into 4 separate choices. 

\begin{enumerate}
\item \textbf{Path $\rho_t$.} Diffusion/score methods often define $\rho_t$ as marginals of a forward noising SDE; flow matching defines $\rho_t$ through chosen probability paths (interpolations or couplings) between endpoints \citep{song2021sde,lipman2022flowmatching,liu2022rectified}.
\item \textbf{Learned object.} Score-based methods learn $s_t=\nabla\log\rho_t$, typically via denoising score matching; flow matching learns $v_t$ directly as a regression target \citep{hyvarinen2005score,vincent2011dsm,lipman2022flowmatching}.
\item \textbf{Sampling dynamics.} Sampling may proceed via reverse-time SDE integration (stochastic sampling) or deterministic ODE integration (probability flow ODE / CNF-style sampling) \citep{song2021sde,chen2018neuralode,grathwohl2019ffjord}.
\item \textbf{Objective weighting and numerical error.} Practical performance depends on how losses weight time or noise levels and on numerical discretization effects in SDE/ODE solvers---issues that become salient when translating continuous-time into finite-step algorithms \citep{ho2020ddpm,song2021sde,kingma2021vdm,karras2022edm}.
\end{enumerate}

\subsection{Contributions of this survey}\label{subsec:contrib}

This survey makes four main contributions.

\begin{enumerate}
\item \textbf{A unified transport viewpoint.}
We present diffusion, score-based, and flow-matching methods within a common framework of probability transport, using SDE/ODE dynamics and their associated PDEs (Fokker--Planck and continuity equations) \citep{song2021sde,lipman2022flowmatching}.

\item \textbf{An equivalence map for samplers.}
We clarify the relationship between reverse-time SDE sampling and deterministic probability-flow ODE sampling, identifying the sense in which they share the same one-time marginals and where they differ in path-space behavior and numerical properties \citep{song2021sde}.

\item \textbf{A unified view of training objectives.}
We relate DDPM-style losses, denoising score matching, and continuous-time score-SDE objectives as weighted score-regression problems, and connect these to flow matching as velocity regression under prescribed probability paths \citep{ho2020ddpm,vincent2011dsm,kingma2021vdm,lipman2022flowmatching}.

\item \textbf{A theory--practice bridge.}
We organize approximation, estimation, discretization, and path-mismatch effects into a common error decomposition, and use this perspective to highlight open problems in path design, robustness, conditioning, and fast sampling.
\end{enumerate}

\subsection{Reading guide: three tracks}\label{subsec:reading}

The survey is designed for three complementary reading styles.
\begin{itemize}
\item \textbf{Applied ML track (implementation-first).} Read Section~2 for notation, then prioritize the sampler and comparison sections: reverse-time SDE versus probability flow ODE sampling, flow matching training, and practical tradeoffs such as solver choice, stability, and cost. The ODE/CNF literature is particularly relevant for understanding likelihood computation and numerical sensitivity \citep{chen2018neuralode,grathwohl2019ffjord,song2021sde}.
\item \textbf{ML theory track (equivalences + error decomposition).} Read the full main text, focusing on the boxed propositions connecting (i) time reversal and scores, (ii) SDE/ODE marginal equivalence, and (iii) objective equivalences (score matching / denoising score matching / diffusion losses) and their weighting \citep{song2021sde,kingma2021vdm}.
\item \textbf{Math/stat track (PDE/SDE foundations).} Read Section~2 for notation, then consult the appendices for formal time-reversal results and PDE derivations. Classical reverse-time diffusion modeling provides a rigorous foundation for the reverse-time SDE in modern score-based generative modeling\citep{anderson1982reverse}.
\end{itemize}

\subsection{Organization of the paper}\label{subsec:org}

Section~2 introduces unified notation and collects preliminaries on SDEs/ODEs and their associated PDEs. Sections~3--5 present diffusion and score-based modeling (forward noising, reverse-time sampling, and training objectives). Section~6 discusses probability flow ODEs and their relationship to CNFs and likelihood computation \citep{song2021sde,chen2018neuralode,grathwohl2019ffjord}. Section~7 presents flow matching and clarifies its relationship to diffusion paths and score-based approaches \citep{lipman2022flowmatching}. Section~8 synthesizes comparisons, and Section~9 discusses theory and error sources. Section~10 concludes with open problems, with appendices providing full derivations and more details.

%% file: figures/fig_unified_transport.tex
\begin{figure}[t]
\centering
\resizebox{\textwidth}{!}{%
\begin{tikzpicture}[
    >=Latex,
    font=\footnotesize,
    node distance=1.3cm and 1.6cm,
    dist/.style={
        draw,
        circle,
        minimum size=1.2cm,
        align=center
    },
    box/.style={
        draw,
        rounded corners,
        align=center,
        inner sep=5pt,
        text width=3.3cm
    },
    smallbox/.style={
        draw,
        rounded corners,
        align=center,
        inner sep=4pt
    },
    labelbox/.style={
        align=center,
        inner sep=2pt
    },
    every path/.style={line width=0.8pt}
]

\node[dist] (rho0) {$\rho_0$\\data};
\node[dist, right=3.8cm of rho0] (rhot) {$\rho_t$\\intermediate};
\node[dist, right=3.8cm of rhot] (rho1) {$\rho_1$\\noise};

\draw[->] (rho0) -- node[above=4pt, fill=white, inner sep=1pt] {forward path} (rhot);
\draw[->] (rhot) -- (rho1);

\node[labelbox, above=0.7cm of rhot] {\textbf{Probability transport view}};
\node[labelbox, below=0.65cm of rhot] {transport between data and reference};

\node[box, below=1.7cm of rho0] (fsde) {forward SDE\\$dX_t=f(X_t,t)\,dt+g(t)\,dW_t$};
\node[box, right=4.1cm of fsde] (score) {learn score field\\$s_\theta(x,t)\approx \nabla_x\log \rho_t(x)$};
\node[box, right=4.1cm of score] (rsde) {reverse-time SDE\\sampling};

\draw[->] (fsde) -- node[above=4pt, fill=white, inner sep=1pt] {same marginals} (score);
\draw[->] (score) -- node[above=4pt, fill=white, inner sep=1pt] {reverse sampling} (rsde);

\node[labelbox, left=0.4cm of fsde] {\textbf{Diffusion / score view}};

\draw[dashed, ->] (fsde.north) to[out=80,in=-90] (rho0.south);
\draw[dashed, ->] (score.north) to[out=90,in=-90] (rhot.south);
\draw[dashed, ->] (rsde.north) to[out=100,in=-90] (rho1.south);

\node[box, below=2.0cm of fsde] (pfode) {probability-flow ODE\\$v_{\mathrm{PF}}(x,t)=f(x,t)-\tfrac12 g(t)^2 s_t(x)$};
\node[box, right=4.1cm of pfode] (same) {same one-time marginals\\as forward SDE};
\node[box, right=4.1cm of same] (fm) {flow matching\\learn velocity $v_\theta(x,t)$};

\draw[->] (pfode) -- node[above=4pt, fill=white, inner sep=1pt] {ODE transport} (same);
\draw[->] (same) -- node[above=4pt, fill=white, inner sep=1pt] {direct velocity learning} (fm);

\node[labelbox, left=0.4cm of pfode] {\textbf{Deterministic transport view}};

\draw[dashed,->]
(score.east) -- ++(0.9,0)
             |- node[pos=0.25, right, fill=white, inner sep=1pt] {\scriptsize induce velocity}
                (pfode.north east);

\draw[dashed,->]
(fm.north) -- ++(0,0.7)
          -| (rhot.east);

\node[smallbox, below=1.5cm of same, text width=11.5cm] (summary) {
\textbf{Main design choices:}
\quad
(1) \emph{path choice} $(\rho_t)$,
\quad
(2) \emph{learned field} (score $s_t$ versus velocity $v_t$),
\quad
(3) \emph{sampling dynamics} (reverse-time SDE versus deterministic ODE)
};

\node[labelbox, below=0.7cm of summary] {\footnotesize
Diffusion learns a score field and can sample via either a reverse-time SDE or the associated probability-flow ODE; flow matching specifies a path directly and learns the corresponding velocity field.
};

\end{tikzpicture}%
}
\caption{Unified view of diffusion, score-based, and flow-matching generative models as probability transport between a data distribution $\rho_0$ and a reference distribution $\rho_1$.}
\label{fig:unified_transport}
\end{figure}

%% file: sections/section2_preliminaries.tex
This section fixes notation and collects the dynamical identities reused throughout the survey. Our goal is to state the modeling objects and governing equations in a way that can accommodate (i) discrete-time diffusion models, such as DDPM \citep{ho2020ddpm}, (ii) continuous-time score-based modeling via SDEs \citep{song2021sde}, and (iii) deterministic transport formulations, such as continuous normalizing flows and flow matching \citep{chen2018neuralode,lipman2022flowmatching}. Technical conditions (existence and uniqueness of solutions, smoothness of densities, and boundary behavior) are deferred to the appendices.

\subsection{Time convention, endpoint distributions, and marginals}\label{subsec:time}

We use time $t\in[0,1]$. Unless explicitly stated otherwise, we adopt the \emph{forward} convention
\[
t=0:\ \rho_0 \ \text{(data)}, \qquad t=1:\ \rho_1 \ \text{(noise/prior)},
\]
and generation runs in reverse time from $\rho_1$ to $\rho_0$. Let $X_t\in\mathbb{R}^d$ denote a time-indexed random variable. We write $\mu_t$ for its probability law and $\rho_t$ for its density when that density exists. In particular, $\mu_0\equiv \mu_{\mathrm{data}}$ and, when convenient, $\rho_0\equiv \rho_{\mathrm{data}}$. The terminal distribution $\mu_1$ is taken to be a tractable reference law, typically $\mathcal{N}(0,I)$.

Two modeling choices recur throughout the paper:
\begin{enumerate}
\item \textbf{The probability path} $(\mu_t)_{t\in[0,1]}$: how intermediate marginals are defined (e.g., as SDE marginals in diffusion/score models, or via an explicit interpolation/coupling in flow matching).
\item \textbf{The learned field}: whether we learn a \emph{score field} $s_t(x)=\nabla_x\log\rho_t(x)$ or a \emph{velocity field} $v_t(x)$ that transports mass along the path.
\end{enumerate}

Throughout the main text we use density notation when it is available, but many statements are more fundamentally statements about probability measures. For example, when we later say that two dynamics have the ``same marginals,'' this should be understood as equality of the one-time laws $\mu_t$ for each $t$.

\subsection{Discrete-time diffusion notation}\label{subsec:ddpm_notation}

In discrete-time diffusion models, one often writes a forward noising chain
\[
q(x_t\mid x_{t-1}) = \mathcal{N}\!\big(\sqrt{1-\beta_t}\,x_{t-1},\,\beta_t I\big),
\]
with variance schedule $(\beta_t)_{t=1}^T$ \citep{ho2020ddpm}. It is standard to define
\[
\alpha_t = 1-\beta_t, \qquad \bar\alpha_t = \prod_{s=1}^t \alpha_s,
\]
so that the perturbed sample admits the closed-form reparameterization
\[
x_t = \sqrt{\bar\alpha_t}\,x_0 + \sqrt{1-\bar\alpha_t}\,\varepsilon,
\qquad \varepsilon\sim\mathcal{N}(0,I).
\]
These discrete-time quantities will reappear in later sections when we compare DDPM training objectives with their continuous-time counterparts.

\subsection{Forward diffusions as SDEs and the Fokker--Planck equation}\label{subsec:sde}

A forward diffusion is commonly expressed as an It\^{o} SDE on $\mathbb{R}^d$,
\begin{equation}\label{eq:forward_sde}
dX_t = f(X_t,t)\,dt + g(t)\,dW_t,
\end{equation}
where $f:\mathbb{R}^d\times[0,1]\to\mathbb{R}^d$ is a drift field, $g:[0,1]\to\mathbb{R}_+$ is a scalar diffusion coefficient (extensions to matrix-valued diffusion are standard), and $(W_t)_{t\in[0,1]}$ is a standard Wiener process. When $\mu_t$ admits a density $\rho_t$, the latter satisfies the forward Kolmogorov (Fokker--Planck) equation
\begin{equation}\label{eq:fp}
\partial_t \rho_t(x) = -\nabla\cdot\!\big(\rho_t(x)\, f(x,t)\big) + \tfrac12 g(t)^2\,\Delta \rho_t(x).
\end{equation}
Diffusion modeling typically chooses $(f,g)$ so that if $X_0\sim\mu_0$, then $\mu_1$ is close to a known reference law, often Gaussian, enabling generation by reversing the dynamics \citep{ho2020ddpm,song2021sde}.

\subsection{Deterministic transport as ODEs and the continuity equation}\label{subsec:ode}

A deterministic transport model evolves particles via an ODE,
\begin{equation}\label{eq:forward_ode}
\frac{dX_t}{dt} = v(X_t,t),
\end{equation}
where $v:\mathbb{R}^d\times[0,1]\to\mathbb{R}^d$ is a velocity field. If $X_0\sim\mu_0$ and the flow map is well-defined, the induced marginals satisfy the continuity equation
\begin{equation}\label{eq:continuity}
\partial_t \rho_t(x) + \nabla\cdot\!\big(\rho_t(x)\, v(x,t)\big)=0
\end{equation}
whenever $\mu_t$ admits a density $\rho_t$. Continuous normalizing flows use \eqref{eq:forward_ode} to construct invertible transport between endpoints while enabling likelihood evaluation through instantaneous change-of-variables \citep{chen2018neuralode}. In particular, along a trajectory $X_t$ one has
\begin{equation}\label{eq:icov}
\frac{d}{dt}\log \rho_t(X_t) = -\nabla\cdot v(X_t,t).
\end{equation}
Flow matching can be viewed as learning a velocity field $v_\theta$ so that \eqref{eq:continuity} holds along a prescribed probability path \citep{lipman2022flowmatching}.

\subsection{Score functions and score models}\label{subsec:score}

When $\mu_t$ admits a differentiable and strictly positive density $\rho_t$, the \emph{score} is defined by
\begin{equation}\label{eq:score_def}
s_t(x) \;=\; \nabla_x \log \rho_t(x).
\end{equation}
Score-based generative modeling learns a neural approximation $s_\theta(x,t)\approx s_t(x)$ for a range of noise levels (times) and uses it to define reverse-time sampling dynamics \citep{song2021sde}. In practice, the score is learned via regression losses derived from score matching and denoising score matching \citep{hyvarinen2005score,vincent2011dsm}, leveraging tractable perturbation kernels that relate clean samples $x_0\sim\mu_0$ to noisy samples $x_t$.

The dependence of score-based methods on densities is worth emphasizing: score notation is available only when the intermediate laws admit sufficiently regular densities. This is one reason diffusive perturbations are convenient: they tend to regularize the distribution enough that density-based objects become well-defined for positive times.

\subsection{Fields on probability space: score vs.\ velocity}\label{subsec:fields}

Many modern generative methods can be described by learning one of two field types:
\begin{enumerate}
\item \textbf{Score fields} $s_\theta(x,t)$, which enter reverse-time SDE samplers and the deterministic probability-flow ODE derived from the same forward diffusion \citep{song2021sde}.
\item \textbf{Velocity fields} $v_\theta(x,t)$, which directly specify deterministic transport via \eqref{eq:forward_ode}--\eqref{eq:continuity} and are learned in flow matching as regression targets under a path distribution \citep{lipman2022flowmatching}.
\end{enumerate}
Later sections make precise when these parameterizations coincide (e.g., via the probability-flow construction for a given forward diffusion) and when they differ (e.g., because flow matching permits broader path choices and training targets).

\subsection{Notations and Symbols}\label{subsec:cheatsheet}

Table~\ref{tab:notation} summarizes the principal symbols used throughout the paper.

\begin{table}[t]
\centering
\begin{tabular}{ll}
\hline
Symbol & Meaning \\
\hline
$\mu_0$ & Data distribution (probability measure) \\
$\rho_0$ & Data density, when it exists \\
$\mu_1$ & Reference/noise distribution (often $\mathcal{N}(0,I)$) \\
$\mu_t,\rho_t$ & Intermediate law / density at time $t\in[0,1]$ \\
$X_t$ & Random variable with law $\mu_t$ \\
$f(x,t),\,g(t)$ & Drift and diffusion coefficients of forward SDE \eqref{eq:forward_sde} \\
$W_t$ & Wiener process in \eqref{eq:forward_sde} \\
$s_t(x)$ & Score $\nabla_x\log\rho_t(x)$ \eqref{eq:score_def} \\
$s_\theta(x,t)$ & Learned score model \\
$v(x,t)$ & Velocity field of ODE \eqref{eq:forward_ode} \\
$v_\theta(x,t)$ & Learned velocity model (flow matching / CNF) \\
$\beta_t,\alpha_t,\bar\alpha_t$ & Standard discrete-time diffusion schedule parameters \\
\hline
\end{tabular}
\caption{Unified notation used throughout the survey.}
\label{tab:notation}
\end{table}

%% file: sections/section3_forward_process.tex
Diffusion and score-based generative models specify a \emph{forward} corruption process that maps data $x_0\sim \mu_0$ to a tractable reference distribution $\mu_1$ (typically Gaussian). This process induces a probability path $(\mu_t)_{t\in[0,1]}$, or densities $(\rho_t)_{t\in[0,1]}$ when they exist, that is later traversed in reverse for generation. We summarize the two most common constructions: (i) discrete-time Markov chains (DDPM-style) \citep{ho2020ddpm,nichol2021improved} and (ii) continuous-time SDEs (score-SDE view) \citep{song2021sde}. The objects that recur in training, namely, the marginal perturbation kernel and the resulting time-dependent signal-to-noise ratio. Recent work shows that this design space extends well beyond standard Gaussian forward noising, including structured discrete-state denoising, continuous-time jump-process corruptions, and physics-inspired alternatives to diffusion transport \citep{austin2021d3pm,campbell2022ctddm,xu2022pfgm,xu2023pfgmpp}.

\subsection{Discrete-time forward noising (DDPM)}\label{subsec:ddpm_forward}

In discrete-time diffusion models, the forward process can be described as a Markov chain $(x_t)_{t=0}^T$ with Gaussian transitions
\begin{equation}\label{eq:ddpm_forward}
q(x_t\mid x_{t-1}) = \mathcal{N}\!\big(\sqrt{1-\beta_t}\,x_{t-1},\ \beta_t I\big),
\end{equation}
where $(\beta_t)_{t=1}^T$ is a variance schedule \citep{ho2020ddpm}. This choice yields a closed-form marginal perturbation kernel
\begin{equation}\label{eq:q_xt_x0}
q(x_t\mid x_0)=\mathcal{N}\!\big(\sqrt{\bar\alpha_t}\,x_0,\ (1-\bar\alpha_t)I\big),
\qquad \bar\alpha_t=\prod_{s=1}^t (1-\beta_s),
\end{equation}
so that a noisy sample can be generated as
\begin{equation}\label{eq:ddpm_reparam}
x_t = \sqrt{\bar\alpha_t}\,x_0 + \sqrt{1-\bar\alpha_t}\,\varepsilon,
\qquad \varepsilon\sim\mathcal{N}(0,I).
\end{equation}
The schedule $(\beta_t)$ determines how quickly information about $x_0$ is erased and therefore controls both training difficulty and the numerical behavior of reverse-time sampling. Practical improvements often modify the schedule or the parameterization of the reverse model while preserving the same basic forward kernel \citep{nichol2021improved}.

Although \eqref{eq:ddpm_forward} uses Gaussian transitions, the same denoising template has been extended to continuous space. D3PMs replace Gaussian noising by structured discrete-state transition kernels \citep{austin2021d3pm}, while star-shaped diffusion models alter the geometry of the forward process itself \citep{okhotin2023ssddpm}. These variants reinforce a central point of this survey: the forward process should be regarded as a modeling choice, not as a fixed Gaussian recipe.

\subsection{Continuous-time forward noising (score-SDE view)}\label{subsec:sde_forward}

A continuous-time forward process is specified by an SDE of the form
\[
dX_t = f(X_t,t)\,dt + g(t)\,dW_t,\qquad t\in[0,1],
\]
chosen so that $X_0\sim \mu_0$ and $X_1\sim \mu_1$ for a tractable reference law $\mu_1$ \citep{song2021sde}. Two widely used families are:

\paragraph{Variance-preserving (VP) SDE.}
A common choice is
\begin{equation}\label{eq:vp_sde}
dX_t = -\tfrac12 \beta(t)\,X_t\,dt + \sqrt{\beta(t)}\,dW_t,
\end{equation}
where $\beta(t)>0$ is a continuous noise schedule. The corresponding conditional marginals remain Gaussian:
\begin{equation}\label{eq:vp_marginal}
X_t \mid X_0=x_0 \sim \mathcal{N}\!\big(\alpha(t)\,x_0,\ \sigma(t)^2 I\big),
\end{equation}
with $\alpha(t)$ and $\sigma(t)$ determined by $\beta(t)$ \citep{song2021sde}. This family can be viewed as the continuous-time limit of DDPM-style chains under appropriate scaling.

\paragraph{Variance-exploding (VE) SDE.}
Another common choice is
\begin{equation}\label{eq:ve_sde}
dX_t = \sqrt{\frac{d\,\sigma(t)^2}{dt}}\,dW_t,
\end{equation}
which keeps the mean fixed while increasing the noise level from $\sigma(0)\approx 0$ to a large $\sigma(1)$. Again, $X_t\mid X_0=x_0$ is Gaussian with mean $x_0$ and variance $\sigma(t)^2 I$ \citep{song2021sde}. VE processes are often convenient for score estimation across a wide range of noise scales.

Other variants, such as sub-VP SDEs and critically damped Langevin diffusions, adjust the drift and diffusion to trade off likelihood estimation properties and sampling behavior \citep{song2021sde,dockhorn2022cld}. Even though these alternative methods might employ certain constraints to modify their likelihood estimation, the primary premise of the forward  process remains an instrumental feature.

\subsection{Perturbation kernels and reparameterizations}\label{subsec:kernels}

Most training objectives for diffusion and score-based models reduce to expectations over \emph{perturbed} samples $x_t$ together with the time index $t$. The central analytical convenience is that many forward processes yield Gaussian perturbation kernels of the form
\begin{equation}\label{eq:gaussian_kernel}
x_t = m(t)\,x_0 + s(t)\,\varepsilon,\qquad \varepsilon\sim\mathcal{N}(0,I),
\end{equation}
for scalar functions $m(t)$ (signal coefficient) and $s(t)$ (noise scale). For discrete-time DDPMs, $m(t)=\sqrt{\bar\alpha_t}$ and $s(t)=\sqrt{1-\bar\alpha_t}$ \citep{ho2020ddpm}. For VP/VE SDEs, analogous expressions follow from the Gaussian conditional laws above \citep{song2021sde}.

A useful derived quantity is the \emph{signal-to-noise ratio} (SNR), which, up to conventions, scales like $(m(t)/s(t))^2$. Schedules that allocate more training mass to either high-SNR (light noise) or low-SNR (heavy noise) regimes can materially affect sample quality and the stability of reverse-time solvers. Modern analyses of diffusion design often emphasize precisely this schedule/path viewpoint \citep{karras2022edm}.

\subsection{From discrete-time DDPM to continuous-time VP SDE}\label{subsec:ddpm_to_sde}

The discrete-time chain \eqref{eq:ddpm_forward} can be viewed as a time discretization of a continuous-time diffusion. This bridge is useful because it explains why (i) DDPM training objectives admit continuous-time limits and (ii) reverse-time sampling can be derived cleanly via SDE time reversal \citep{song2021sde,sohldickstein2015deep}.

\paragraph{DDPM as a discretization of a VP (Ornstein--Uhlenbeck-type) SDE.}
Consider the DDPM forward transition
\[
x_{k+1} = \sqrt{1-\beta_{k+1}}\,x_k + \sqrt{\beta_{k+1}}\,\varepsilon_{k+1},
\qquad \varepsilon_{k+1}\sim\mathcal{N}(0,I),
\]
with schedule $(\beta_k)_{k=1}^T$ \citep{ho2020ddpm}. Introduce a time step $\Delta t = 1/T$ and associate $t_k = k\Delta t$. Suppose
\[
\beta_{k+1} = \beta(t_k)\,\Delta t + o(\Delta t),
\]
for some smooth function $\beta(t)>0$. Using the Taylor approximation $\sqrt{1-u}=1-\tfrac12 u + o(u)$, we obtain
\[
\sqrt{1-\beta_{k+1}}
= 1 - \tfrac12 \beta(t_k)\Delta t + o(\Delta t),
\qquad
\sqrt{\beta_{k+1}}
= \sqrt{\beta(t_k)}\,\sqrt{\Delta t} + o(\sqrt{\Delta t}).
\]
Substituting into the update yields
\[
x_{k+1}-x_k
= -\tfrac12 \beta(t_k)\,x_k\,\Delta t
+ \sqrt{\beta(t_k)}\,\sqrt{\Delta t}\,\varepsilon_{k+1}
+ o(\Delta t).
\]
Since $\sqrt{\Delta t}\,\varepsilon_{k+1}$ has the same distribution as a Brownian increment $\Delta W_k\sim \mathcal{N}(0,\Delta t\,I)$, this is precisely an Euler--Maruyama discretization of the VP SDE
\[
dX_t = -\tfrac12 \beta(t)\,X_t\,dt + \sqrt{\beta(t)}\,dW_t.
\]
In this sense, DDPM forward noising converges to a VP SDE as $T\to\infty$ under the above scaling, and discrete-time diffusion objectives can be interpreted as discretizations of continuous-time score-based objectives.

\subsection{Marginals versus path measures}\label{subsec:marginals_vs_paths}

A forward diffusion specifies more than the family of one-time marginals $(\mu_t)$: it induces a full \emph{path measure} on trajectories $(X_t)_{t\in[0,1]}$ through the SDE dynamics \citep{song2021sde}. Many equivalence statements used later in the survey should be read with this distinction in mind. For example, the probability-flow ODE associated with a forward SDE is constructed to match the \emph{same marginals} $(\mu_t)$ while generally inducing a different distribution over paths. Conversely, different stochastic processes can share the same endpoint laws but define distinct intermediate marginals and trajectory laws.

This perspective becomes especially important when comparing diffusion and score-based modeling to flow matching: flow matching begins by specifying a coupling/interpolation scheme, and hence a path measure, and then learns a velocity field consistent with that chosen path \citep{lipman2022flowmatching}. Relatedly, Schr\"odinger bridge formulations make the path-measure viewpoint explicit by posing transport as an optimization over stochastic path measures subject to endpoint constraints \citep{debortoli2021dsb}. We return to this distinction when discussing sampler equivalences and conditioning methods in later sections.

\subsection{Probability paths beyond diffusions}\label{subsec:path_beyond}

Flow matching makes the dependence on the probability path explicit: rather than taking $(\mu_t)$ as the marginals of a fixed forward SDE, one may define $(\mu_t)$ through a coupling/interpolation scheme and then learn a velocity field consistent with that path \citep{lipman2022flowmatching}. From this perspective, diffusion models correspond to one important stochastic path family, while alternative paths---for example, the ``straightened'' trajectories emphasized in rectified flow---aim to improve numerical properties such as the effectiveness of coarse discretization \citep{liu2022rectified}.

This broader path-based viewpoint is also useful beyond continuous Gaussian perturbations. For discrete data, continuous-time denoising models define diffusion-like corruption and recovery processes using jump dynamics rather than Gaussian noise \citep{campbell2022ctddm}. Related approaches such as Analog Bits and masked diffusion adapt diffusion-style training to noncontinuous state spaces by embedding or masking discrete structure in ways that preserve tractable learning targets \citep{chen2023analogbits,sahoo2024simplemaskedlm}. Similar ideas have also been explored for discrete-state graph generation \citep{xu2024discretectgraph}. At the same time, physics-inspired alternatives such as Poisson Flow Generative Models and PFGM++ show that the same endpoint-generation problem can be approached through transport constructions that differ substantially from standard Gaussian diffusion while preserving a closely related generative objective \citep{xu2022pfgm,xu2023pfgmpp}. We return to these connections when comparing broader families of generative transports in later sections.

%% file: sections/section4_reverse_dynamics.tex
Given a forward corruption process (Section~\ref{sec:forward}) that maps the data law $\mu_0$ to a tractable reference law $\mu_1$, generation proceeds by approximately simulating a \emph{reverse-time} dynamic that transports samples from $\mu_1$ back to $\mu_0$. For diffusion and score-based models, the reverse-time sampler can be derived from stochastic time reversal and depends on the \emph{score} of the intermediate marginals. This section states the reverse-time SDE and explains how a learned score model gives rise to practical samplers.

\subsection{Reverse-time SDE: where the score enters}\label{subsec:reverse_sde}

Consider the forward It\^{o} diffusion
\[
dX_t = f(X_t,t)\,dt + g(t)\,dW_t,\qquad t\in[0,1],
\]
and assume that its one-time laws admit sufficiently smooth positive densities $\rho_t$. Under suitable regularity assumptions, the time-reversed process satisfies an SDE whose drift contains the score $\nabla_x\log\rho_t(x)$ \citep{anderson1982reverse,song2021sde}. Informally, the reverse-time dynamics can be written as
\begin{equation}\label{eq:reverse_sde_informal}
dX_t = \Big(f(X_t,t) - g(t)^2\,\nabla_x \log \rho_t(X_t)\Big)\,dt + g(t)\,d\bar W_t,
\end{equation}
run from $t=1$ down to $t=0$, where $\bar W_t$ denotes a reverse-time Wiener process.\footnote{Different conventions absorb sign changes into the direction of integration. We follow the common convention of writing the reverse-time SDE with drift evaluated at time $t$ while integrating from $1$ to $0$; see \citet{song2021sde} for a precise formulation.}

\begin{proposition}[Reverse-time SDE and the score]
Let $(X_t)_{t\in[0,1]}$ solve the forward SDE
\[
dX_t = f(X_t,t)\,dt + g(t)\,dW_t,
\]
and suppose the marginals admit sufficiently smooth positive densities $\rho_t$. Then, under standard time-reversal regularity assumptions, the reverse-time dynamics are given by an SDE whose drift contains the score of the forward marginals:
\[
dX_t = \Big(f(X_t,t)-g(t)^2 \nabla_x \log \rho_t(X_t)\Big)\,dt + g(t)\,d\bar W_t,
\]
interpreted in reverse time.
\label{prop:reverse_sde}
\end{proposition}

Proposition~\ref{prop:reverse_sde} is the mathematical reason score estimation suffices for generation: once the score field $s_t(x)=\nabla\log\rho_t(x)$ is approximated along the forward path, reverse-time sampling can be defined directly. A full derivation goes back to classical time-reversal results for diffusions \citep{anderson1982reverse}; the score-SDE formulation of \citet{song2021sde} makes this usable in modern generative modeling.

\subsection{Learning the score field}\label{subsec:learn_score}

In practice, $\rho_t$ is unknown, so we learn a neural network $s_\theta(x,t)$ that approximates $s_t(x)$. Training is typically based on denoising score matching: one samples $x_0\sim\mu_0$, chooses a time $t$, generates a perturbed sample $x_t$ from the forward perturbation kernel (Section~\ref{subsec:kernels}), and regresses $s_\theta(x_t,t)$ toward the analytically available conditional score $\nabla_{x_t}\log q(x_t\mid x_0)$ \citep{vincent2011dsm,song2021sde}. Section~5 discusses these objectives and their equivalences in detail. For the purposes of the present section, the key point is that once $s_\theta$ has been trained, it can be substituted for the unknown score in \eqref{eq:reverse_sde_informal}.

\subsection{Practical samplers: discretizing the reverse SDE}\label{subsec:discretize}

To generate samples, one initializes $x_1\sim\mu_1$ (typically $\mathcal{N}(0,I)$) and numerically integrates the reverse-time SDE from $t=1$ to $t=0$ using the learned score $s_\theta$. The simplest discretization is an Euler--Maruyama step of the schematic form
\[
x_{t-\Delta t} = x_t + \Big(f(x_t,t)-g(t)^2 s_\theta(x_t,t)\Big)\Delta t + g(t)\sqrt{\Delta t}\,z,\qquad z\sim\mathcal{N}(0,I).
\]
Exact implementations differ according to time parameterization and sign convention, but the essential structure is the same: a deterministic drift term driven by the learned score together with a stochastic diffusion term. More accurate or stable samplers use higher-order SDE solvers or \emph{predictor--corrector} schemes that alternate a predictor step (SDE discretization) with a corrector step based on Langevin dynamics \citep{song2021sde}. In discrete-time DDPMs, the reverse-time sampler is often implemented as an ancestral reverse Markov chain whose parameters are predicted by a neural network \citep{ho2020ddpm}.

\subsection{SDE versus ODE sampling (preview)}\label{subsec:sde_vs_ode_preview}

The reverse-time SDE sampler is stochastic: even with a fixed initial noise seed, injected noise at each discretization step influences the output. A complementary deterministic alternative is obtained by constructing a \emph{probability flow ODE} whose solution shares the same one-time marginals as the forward SDE \citep{song2021sde}. This ODE viewpoint connects diffusion sampling to continuous normalizing flows and enables likelihood computation via instantaneous change-of-variables, while generally inducing a different path measure from the SDE. We discuss this construction in Section~6.

%% file: sections/section5_objectives.tex
Sections~\ref{sec:forward}--\ref{sec:reverse} described how forward noising processes induce a probability path $(\rho_t)$ and how reverse-time sampling depends on the unknown score field $s_t=\nabla\log\rho_t$. We now describe how modern diffusion/score models \emph{learn} the score (or an equivalent parameterization) from data. The key unifying theme is that most objectives reduce to \emph{weighted regression} of analytically available targets under perturbed data, and many seemingly different losses (DDPM noise prediction, score matching, and continuous-time objectives) are equivalent up to time-dependent weights and parameterizations \citep{ho2020ddpm,vincent2011dsm,song2021sde}.

\subsection{Score matching and denoising score matching}\label{subsec:sm_dsm}

Classical score matching estimates the score of an unknown density by minimizing the Fisher divergence between a model score and the data score \citep{hyvarinen2005score}. In generative modeling, the most widely used variant is \emph{denoising score matching} (DSM), which leverages a tractable corruption kernel $q_\sigma(x\mid x_0)$ (often Gaussian) and the identity
\[
\nabla_x \log q_\sigma(x) \;=\; \mathbb{E}\big[\nabla_x \log q_\sigma(x\mid x_0)\mid x\big],
\]
to train a score network without ever evaluating $q_\sigma(x)$ directly \citep{vincent2011dsm}. In the diffusion setting, the ``noise level'' $\sigma$ is replaced by time $t$, and DSM is applied to the family of perturbed marginals $\rho_t$, induced by the forward process \citep{song2021sde}.

For Gaussian perturbations of the form $x_t = m(t)x_0 + s(t)\varepsilon$ (Section~\ref{subsec:kernels}), the conditional density $q(x_t\mid x_0)$ is Gaussian and its score with respect to $x_t$ is available in closed form:
\begin{equation}\label{eq:cond_score_gauss}
\nabla_{x_t}\log q(x_t\mid x_0) \;=\; -\frac{1}{s(t)^2}\Big(x_t - m(t)\,x_0\Big).
\end{equation}
DSM then regresses the model score $s_\theta(x_t,t)$ toward \eqref{eq:cond_score_gauss} in expectation over $(x_0,t,\varepsilon)$.

\subsection{DDPM training as noise prediction (and score prediction)}\label{subsec:ddpm_noise}

DDPMs are often trained by predicting the noise $\varepsilon$ in the reparameterization \eqref{eq:ddpm_reparam}:
\begin{equation}\label{eq:ddpm_eps_loss}
\mathcal{L}_{\varepsilon}(\theta)
=
\mathbb{E}_{t,x_0,\varepsilon}\Big[
\big\|\varepsilon - \varepsilon_\theta(x_t,t)\big\|^2
\Big],
\qquad x_t=\sqrt{\bar\alpha_t}x_0+\sqrt{1-\bar\alpha_t}\varepsilon,
\end{equation}
possibly with time-dependent weights \citep{ho2020ddpm}. This objective is equivalent to score regression under a change of variables: for the Gaussian kernel \eqref{eq:q_xt_x0}, the conditional score \eqref{eq:cond_score_gauss} can be expressed in terms of $\varepsilon$, and a predictor $\varepsilon_\theta$ induces a score model via
\begin{equation}\label{eq:eps_to_score}
s_\theta(x_t,t)
\;\approx\;
-\frac{1}{\sqrt{1-\bar\alpha_t}}\,\varepsilon_\theta(x_t,t)
\quad \text{(up to known scalings).}
\end{equation}
Thus ``noise prediction'' and ``score prediction'' are largely different parameterizations of the same regression problem, with different implicit weightings across time/noise levels. Improved DDPM variants discuss alternative parameterizations and weighting choices \citep{nichol2021improved}.

\subsection{Continuous-time objectives and weighting}\label{subsec:ct_objectives}

In the continuous-time score-SDE framework, one typically samples $t\sim \mathrm{Unif}[0,1]$ (or another distribution) and minimizes a time-weighted DSM objective of the form
\begin{equation}\label{eq:ct_dsm}
\mathcal{L}(\theta)
=
\mathbb{E}_{t}\,\mathbb{E}_{x_0,\varepsilon}\Big[
\lambda(t)\,\big\| s_\theta(x_t,t) - \nabla_{x_t}\log q(x_t\mid x_0)\big\|^2
\Big],
\end{equation}
where $\lambda(t)$ is a weighting function \citep{song2021sde}. Different choices of $\lambda(t)$ correspond to emphasizing different SNR regimes along the path, and can strongly affect both sample quality and the stiffness/discretization sensitivity of reverse-time solvers (cf.\ the schedule/path viewpoint in \citep{karras2022edm}).

\subsection{Equivalence view: weighted Fisher divergence (informal)}\label{subsec:fisher_informal}

Many diffusion/score objectives can be interpreted as minimizing a \emph{time-integrated Fisher divergence} between the learned score and the true score of $\rho_t$ (up to constants and weighting). Informally,
\[
\mathcal{L}(\theta) \ \approx\ \int_0^1 \lambda(t)\, \mathbb{E}_{x\sim \rho_t}\big[\|s_\theta(x,t)-s_t(x)\|^2\big]\,dt,
\]
where the relationship between $\lambda(t)$ and the forward process depends on the chosen parameterization and discretization \citep{ho2020ddpm,song2021sde,kingma2021vdm}. We defer precise equivalence statements (including the connection to ELBO-style objectives for discrete-time diffusion models) to Appendix~C and focus here on the practical implications: \emph{different ``losses'' often differ primarily by the time/noise weighting they induce.}

\begin{proposition}[Diffusion objectives as weighted score regression]
\label{prop:weighted_fisher}
For Gaussian perturbation kernels and the usual diffusion parameterizations, the standard training objectives used in DDPMs, denoising score matching, and continuous-time score-SDE training can all be written, up to constants and schedule-dependent weights, as minimizing a time-integrated squared error between a model score and the true score of the perturbed marginals:
\[
\int_0^1 \lambda(t)\,\mathbb{E}_{x\sim\rho_t}\!\left[\|s_\theta(x,t)-s_t(x)\|^2\right]\,dt.
\]
Equivalently, common noise-prediction objectives are reparameterized score-regression objectives under Gaussian perturbations.
\end{proposition}

Proposition~\ref{prop:weighted_fisher} formalizes the unifying claim of this subsection: DDPM noise-prediction losses, denoising score matching, and continuous-time score-SDE objectives differ mainly in parameterization and weighting, rather than in the statistical object they estimate. In the main text we keep this statement informal; the derivational details are deferred to Appendix~C.
\subsection{Practice note: parameterizations and what changes}\label{subsec:param_practice}

A recurring source of confusion is that implementations may train networks to predict different quantities---$\varepsilon$, $x_0$, the mean of $p_\theta(x_{t-1}\mid x_t)$, or the score---while describing them as different methods. For Gaussian forward kernels, these parameterizations are algebraically linked by known scalings (e.g., \eqref{eq:eps_to_score}), but the induced optimization problem can change due to weighting and normalization. From a unifying viewpoint, the question is not ``which variable is predicted?'' but rather: \emph{what score/velocity field is implied by the parameterization and what time-dependent weighting does training apply along the path?}

%% file: sections/section6_probability_flow_ode.tex
The reverse-time sampler in Section~\ref{sec:reverse} is stochastic because it integrates an SDE whose drift depends on the score. A complementary deterministic viewpoint---central to unifying diffusion models with continuous normalizing flows---is that the same family of one-time marginals can be generated by an ODE whose vector field also depends on the score. This \emph{probability flow ODE} enables deterministic sampling and, in principle, likelihood computation via instantaneous change-of-variables \citep{song2021sde,chen2018neuralode,grathwohl2019ffjord}.

\subsection{From Fokker--Planck to a deterministic flow}\label{subsec:pfode_derivation}

Consider the forward SDE
\[
dX_t = f(X_t,t)\,dt + g(t)\,dW_t,
\]
and suppose its one-time laws admit densities $\rho_t$ satisfying the Fokker--Planck equation \eqref{eq:fp}. \citet{song2021sde} observed that one can construct a deterministic ODE
\begin{equation}\label{eq:pfode}
\frac{dX_t}{dt} = f(X_t,t) - \tfrac12 g(t)^2\,\nabla_x \log \rho_t(X_t),
\end{equation}
whose induced density evolution matches the same Fokker--Planck marginals. In other words, integrating \eqref{eq:pfode} from $t=0$ to $t=1$ produces the same family of one-time marginals as the stochastic forward SDE (and likewise in reverse time for generation), even though the path measures differ (Section~\ref{subsec:marginals_vs_paths}). This marginal-equivalence property is the formal bridge between diffusion models and deterministic transport.

\begin{proposition}[Probability-flow ODE has the same marginals]
\label{prop:pfode}
Assume the forward SDE marginals admit sufficiently smooth densities $\rho_t$ so that the score $s_t(x)=\nabla_x\log\rho_t(x)$ is well-defined. Then the ODE
\[
\frac{dX_t}{dt} = f(X_t,t) - \tfrac12 g(t)^2\, s_t(X_t)
\]
induces the same one-time marginals as the forward SDE, and hence the same densities $(\rho_t)_{t\in[0,1]}$ whenever those densities exist.
\end{proposition}

The proof is obtained by matching the PDEs governing density evolution: substituting the ODE velocity into the continuity equation reproduces the same Fokker--Planck evolution as the SDE \citep{song2021sde}. Proposition~\ref{prop:pfode} should be read as a \emph{marginal} equivalence statement; the ODE and SDE generally induce different path measures.

In practice, $\nabla_x\log\rho_t$ is replaced by the learned score network $s_\theta(x,t)$, yielding the approximate probability flow ODE
\begin{equation}\label{eq:pfode_theta}
\frac{dX_t}{dt} = f(X_t,t) - \tfrac12 g(t)^2\,s_\theta(X_t,t).
\end{equation}
Sampling is then performed by solving \eqref{eq:pfode_theta} backward in time from $t=1$ to $t=0$ with an ODE solver.

\subsection{Deterministic sampling and numerical considerations}\label{subsec:pfode_sampling}

Compared to SDE sampling, ODE sampling has two practical consequences.

\paragraph{(i) Determinism.}
Given an initial draw $x_1\sim\mu_1$, the ODE trajectory is deterministic up to numerical solver tolerances. This can be advantageous for reproducibility and for downstream tasks where one wants a deterministic mapping from a latent seed to a sample.

\paragraph{(ii) Solver error and stiffness.}
ODE-based sampling replaces stochastic discretization error with deterministic solver error. In high dimensions, stiffness can arise near endpoint regions where the effective score magnitude changes rapidly, making coarse step sizes unstable or biased. This is one reason step-size schedules and solver choices matter in practice; diffusion-design analyses often emphasize the interaction between noise schedules, SNR allocation, and discretization error \citep{karras2022edm}. When adaptive solvers are used, computational cost is controlled indirectly by tolerance parameters rather than by an explicit step count.

\subsection{Likelihood computation via instantaneous change-of-variables}\label{subsec:pfode_likelihood}

Because \eqref{eq:pfode_theta} defines an invertible flow under standard regularity conditions, one can in principle compute log-likelihoods using the instantaneous change-of-variables formula for ODE flows \citep{chen2018neuralode}. Concretely, if $X_t$ follows \eqref{eq:pfode_theta}, then
\begin{equation}\label{eq:pfode_logp}
\frac{d}{dt}\log \rho_t(X_t)
=
-\nabla\cdot \Big(f(X_t,t)-\tfrac12 g(t)^2 s_\theta(X_t,t)\Big),
\end{equation}
and integrating \eqref{eq:pfode_logp} along trajectories relates $\log \rho_0$ and $\log \rho_1$. Computing the divergence term exactly is expensive in high dimensions, and scalable CNF implementations use stochastic trace estimators (e.g., Hutchinson estimators), as in FFJORD \citep{grathwohl2019ffjord}. In diffusion models, this likelihood connection is typically only approximate because both the score model and the numerical solution are approximate. Nonetheless, the probability-flow viewpoint provides a principled bridge between diffusion samplers and likelihood-based continuous flows \citep{song2021sde}.

\subsection{How ODE and SDE samplers relate (summary)}\label{subsec:ode_sde_summary}

Both reverse-time SDE sampling (Section~\ref{sec:reverse}) and probability-flow ODE sampling \eqref{eq:pfode_theta} use the same learned score field. In the idealized setting of an exact score and exact numerical integration, they produce samples from the same target distribution via matching one-time marginals, but they differ in (i) the distribution over intermediate trajectories and (ii) how numerical error manifests. This distinction becomes especially important when comparing diffusion to flow matching: diffusion learns a score field tied to a stochastic forward process and then induces an ODE velocity through \eqref{eq:pfode_theta}, whereas flow matching learns the velocity field directly along a chosen path \citep{song2021sde,lipman2022flowmatching}.

%% file: sections/section7_flow_matching.tex
Flow matching provides a complementary route to generative modeling that emphasizes \emph{deterministic transport} and makes the choice of probability path explicit. Rather than defining $(\mu_t)$ as the marginals of a fixed forward SDE and learning a score field, flow matching specifies a family of intermediate distributions---typically through a coupling and an interpolation rule---and learns a \emph{velocity field} whose induced ODE transports mass along that path \citep{lipman2022flowmatching}. This viewpoint connects naturally to continuous normalizing flows and helps explain recent progress on few-step generation via path straightening, as in rectified flow \citep{liu2022rectified}. More recent work also shows that the same path-based transport principle extends beyond the standard continuous Euclidean setting, including discrete-state, graph-valued, function-valued, and Wasserstein-space formulations \citep{gat2024dfm,cheng2024sfm,davis2024fisherflow,eijkelboom2024vfm,kerrigan2024ffm,haviv2025wfm}.

\subsection{Path-based formulation and conditional velocities}\label{subsec:flowmatching_path}

Let $\pi(x_0,x_1)$ be a coupling between data $x_0\sim \mu_0$ and noise $x_1\sim \mu_1$. A \emph{path sampler} specifies a random intermediate state $x_t$ given $(x_0,x_1,t)$, for example by an affine interpolation
\begin{equation}\label{eq:affine_path}
x_t = a(t)\,x_0 + b(t)\,x_1,
\end{equation}
with scalar functions $a(t)$ and $b(t)$. The induced law of $x_t$ defines a probability path $(\mu_t)_{t\in[0,1]}$, with densities $(\rho_t)$ when they exist.

If the interpolation is differentiable in $t$, then along a sampled endpoint pair $(x_0,x_1)$ the path derivative is
\begin{equation}\label{eq:conditional_velocity}
\dot x_t := \frac{d}{dt}x_t = a'(t)\,x_0 + b'(t)\,x_1.
\end{equation}
This object is naturally a \emph{conditional} target: it depends on the sampled endpoints $(x_0,x_1)$, not only on the current state $x_t$.

By contrast, the velocity field that appears in the continuity equation
\[
\partial_t \rho_t + \nabla\cdot(\rho_t v_t)=0
\]
is a function of the current state and time. For a given path construction, the marginally correct velocity is
\begin{equation}\label{eq:marginal_velocity}
v^*(x,t)=\mathbb{E}\!\left[\dot x_t \mid x_t=x,\,t\right].
\end{equation}
Equation \eqref{eq:marginal_velocity} is the key mathematical object: it is the velocity field whose flow is consistent with the prescribed probability path \citep{lipman2022flowmatching}. In this sense, flow matching can be understood as learning the conditional expectation of the path derivative given the current state.

\begin{proposition}[Optimal flow-matching velocity]
\label{prop:fm_velocity}
Let a path sampler induce intermediate states $x_t$ and conditional path derivative $\dot x_t$. Then the population minimizer of the conditional flow-matching loss
\[
\mathbb{E}\big[\|v_\theta(x_t,t)-\dot x_t\|^2\big]
\]
over measurable functions of $(x_t,t)$ is
\[
v^*(x,t)=\mathbb{E}[\dot x_t\mid x_t=x,\,t].
\]
Equivalently, the optimal unconditional velocity field is the conditional expectation of the path derivative given the current state and time.
\end{proposition}

Proposition~\ref{prop:fm_velocity} is the key mathematical bridge between endpoint-conditioned supervision and the marginal velocity field appearing in the continuity equation. It explains why endpoint-conditioned regression can still recover a state-dependent field suitable for deterministic transport \citep{lipman2022flowmatching}.

\subsection{Conditional and marginal flow matching objectives}\label{subsec:flowmatching_objective}

A practical training objective samples $(x_0,x_1)\sim \pi$, chooses a time $t$, constructs $x_t$, and regresses the model velocity against the conditional target \eqref{eq:conditional_velocity}:
\begin{equation}\label{eq:cfm_loss}
\mathcal{L}_{\mathrm{CFM}}(\theta)
=
\mathbb{E}_{t}\,\mathbb{E}_{(x_0,x_1)\sim \pi}\Big[
\|v_\theta(x_t,t)-\dot x_t\|^2
\Big].
\end{equation}
For the affine path \eqref{eq:affine_path}, this becomes
\[
\mathcal{L}_{\mathrm{CFM}}(\theta)
=
\mathbb{E}_{t}\,\mathbb{E}_{(x_0,x_1)\sim \pi}\Big[
\|v_\theta(x_t,t)-\big(a'(t)x_0+b'(t)x_1\big)\|^2
\Big].
\]

Although \eqref{eq:cfm_loss} is written using endpoint-conditioned targets, its population minimizer is the marginal velocity field \eqref{eq:marginal_velocity}. This follows from the $L^2$ projection principle: minimizing squared error against $\dot x_t$ under the joint law of $(x_t,t,x_0,x_1)$ yields the conditional expectation $\mathbb{E}[\dot x_t\mid x_t,t]$ \citep{lipman2022flowmatching}. This is the basic reason conditional flow matching is mathematically compatible with learning an unconditional field $v_\theta(x,t)$.

An equivalent marginal formulation is therefore
\begin{equation}\label{eq:fm_loss}
\mathcal{L}_{\mathrm{FM}}(\theta)
=
\mathbb{E}_{t}\,\mathbb{E}_{x\sim \rho_t}\Big[
\|v_\theta(x,t)-v^*(x,t)\|^2
\Big],
\end{equation}
where $v^*(x,t)$ is given by \eqref{eq:marginal_velocity}. In practice, the conditional version \eqref{eq:cfm_loss} is preferred because it provides a tractable supervised target via sampled endpoint pairs.

\subsection{How flow matching relates to diffusion and probability-flow ODEs}\label{subsec:flowmatching_relation}

The probability-flow ODE of Section~\ref{sec:pfode} defines a deterministic velocity field
\[
v_{\mathrm{PF}}(x,t)=f(x,t)-\tfrac12 g(t)^2\,s_t(x),
\]
tied to a specific forward diffusion and its score field \citep{song2021sde}. Flow matching can be viewed as learning a velocity field directly, without going through an explicit score parameterization. These viewpoints coincide in special cases:
\begin{itemize}
\item If the chosen path $(\mu_t)$ matches the marginals of a forward diffusion and the target velocity equals the associated probability-flow velocity, then flow matching recovers the same deterministic sampler as the diffusion probability-flow ODE.
\item More generally, flow matching allows broader path families and couplings, potentially yielding velocities that are easier to integrate numerically, less stiff, or better suited to coarse discretization.
\end{itemize}

From the transport viewpoint developed in this survey, diffusion and flow matching differ less in their ultimate goal than in how they parameterize transport: diffusion learns a score field and \emph{induces} a velocity through the probability-flow construction, whereas flow matching learns the velocity field directly along a prescribed path.

\subsection{Gaussian probability paths and tractable supervision}\label{subsec:flowmatching_gaussian_paths}

A major practical advantage of flow matching is that many useful path families admit simple conditional sampling formulas and explicit endpoint-conditioned velocities. In particular, Gaussian probability paths provide tractable supervision analogous to the Gaussian perturbation kernels used in diffusion training \citep{lipman2022flowmatching}. This makes the method scalable: the training loop only needs to sample endpoint pairs, generate an intermediate state $x_t$, and evaluate the explicit conditional target $\dot x_t$.

At the same time, the dependence on the coupling $\pi$ should not be overlooked. Two different couplings with the same endpoint laws can induce different conditional targets and therefore different learned velocity fields. Thus, unlike diffusion, where the forward SDE largely fixes the path, flow matching makes the coupling and interpolation rule explicit modeling choices.

\subsection{Beyond Gaussian paths and Euclidean state spaces}\label{subsec:flowmatching_beyond_euclidean}

Recent work shows that the flow-matching principle extends well beyond finite-dimensional Euclidean state spaces. On the discrete side, discrete flow matching, categorical flow matching on statistical manifolds, and Fisher flow matching all adapt the same basic velocity-learning viewpoint to noncontinuous state spaces \citep{gat2024dfm,cheng2024sfm,davis2024fisherflow}. On structured domains such as graphs, variational flow matching shows that path-based transport ideas can be combined with latent-variable or graph-specific inductive structures \citep{eijkelboom2024vfm}. On infinite-dimensional or nonstandard spaces, functional flow matching and Wasserstein flow matching illustrate that the learned object can be interpreted as a transport field over functions or even over families of probability distributions \citep{kerrigan2024ffm,haviv2025wfm}.

These extensions are conceptually important because they show that flow matching is not merely a particular recipe for images or continuous vectors, but a general strategy for learning transport dynamics in appropriately structured state spaces.

\subsection{Rectified flows and path straightening}\label{subsec:flowmatching_rectified}

Rectified flow proposes to learn transport dynamics whose trajectories are ``as straight as possible,'' with the goal of enabling accurate generation using very few integration steps \citep{liu2022rectified}. From the path-design perspective, rectified flow emphasizes that \emph{choosing or learning a good path} can be as important as the choice between score and velocity parameterizations: straighter paths tend to reduce stiffness and discretization error for coarse solvers, which is crucial for fast sampling.

This viewpoint is conceptually important for the broader comparison developed in the paper. If diffusion highlights the role of score estimation under stochastic noising and probability-flow ODEs highlight marginally equivalent deterministic transport, rectified flow highlights a third axis: the geometry of the path itself.

\subsection{Practice note: what changes when you change the path}\label{subsec:flowmatching_path_practice}

Because flow matching targets the velocity under the path distribution induced by $(\pi,\text{path sampler})$, changing the path changes both the regression target and the distribution over $x_t$ on which the model is trained. This is analogous to changing the noise schedule in diffusion, but more general: the coupling $\pi$ can encode correlations between $x_0$ and $x_1$, and different interpolations can emphasize different geometric aspects of the data manifold.

For downstream tasks such as conditional generation or inverse problems, this path dependence can interact strongly with conditioning mechanisms, since conditioning effectively perturbs the learned velocity or score field along the path. One of the broader lessons of flow matching is therefore that path choice is not merely a technicality: it is a central modeling decision that shapes both training and sampling behavior.

%% file: sections/section8_unified.tex
We now summarize diffusion/score models and flow matching under a common probability-transport lens. The comparisons in this section are intentionally \emph{structural}: they focus on what object is learned (score versus velocity), what path is assumed, what dynamics are solved at sampling time (SDE versus ODE), and where the dominant sources of error arise. These axes are the ones that most directly influence both practice (compute, stability, controllability) and theory (approximation, discretization, and identifiability).

The discussion is anchored by the central formal statements from the preceding sections: Proposition~\ref{prop:reverse_sde} (reverse-time SDE), Proposition~\ref{prop:weighted_fisher} (objective equivalence), Proposition~\ref{prop:pfode} (probability-flow ODE marginal equivalence), and Proposition~\ref{prop:fm_velocity} (optimal flow-matching velocity).

\subsection{At-a-glance taxonomy}\label{subsec:comparison_taxonomy}

Table~\ref{tab:comparison_master} summarizes the main method families through the transport lens developed in this survey: what field is learned, how the path of intermediate distributions is defined, what sampler is used, and where the main numerical or modeling tradeoffs arise.

\begin{table*}[t]
\centering
\small
\begin{tabular}{p{2.4cm}p{2.5cm}p{2.8cm}p{2.4cm}p{2.5cm}p{2.7cm}}
\hline
\textbf{Method family} &
\textbf{Learned object} &
\textbf{Path / intermediate law} &
\textbf{Sampling dynamics} &
\textbf{Likelihood connection} &
\textbf{Main practical / numerical issues} \\
\hline

DDPM / discrete diffusion \citep{ho2020ddpm,nichol2021improved}
&
Noise, mean, or equivalent score parameterization
&
Discrete forward Markov chain with Gaussian noising schedule
&
Reverse ancestral Markov chain
&
ELBO / variational bound; approximate likelihood-related training objective
&
Many denoising steps; schedule sensitivity; discretization bias in the reverse chain \\

Score-SDE models \citep{song2021sde}
&
Score field $s_\theta(x,t)$
&
Forward SDE marginals $(\rho_t)$ induced by VP/VE/sub-VP diffusion
&
Reverse-time SDE
&
Connected to likelihood through the associated probability-flow ODE
&
Stochastic sampling variance; solver choice; endpoint stiffness; score accuracy along the full path \\

Probability-flow ODE \citep{song2021sde,chen2018neuralode,grathwohl2019ffjord}
&
Score field $s_\theta(x,t)$, inducing a velocity field
&
Same one-time marginals as the forward SDE
&
Deterministic ODE
&
Instantaneous change-of-variables; approximate likelihood in principle
&
ODE stiffness; divergence estimation cost; adaptive-solver sensitivity; marginal but not path equivalence \\

Flow matching \citep{lipman2022flowmatching,lipman2024fmguide}
&
Velocity field $v_\theta(x,t)$
&
Explicit coupling/interpolation-defined path
&
Deterministic ODE
&
CNF-style likelihood tools in principle when the flow is well-behaved
&
Path choice strongly affects the learned field; coupling dependence; coarse-step behavior depends on path geometry \\

Rectified flow \citep{liu2022rectified}
&
Velocity field $v_\theta(x,t)$
&
Path straightening / refined coupling intended to make trajectories closer to linear
&
Deterministic ODE
&
Same general CNF-style connection as flow matching
&
Designed for few-step generation; path quality and straightness strongly affect numerical stability \\

\hline
\end{tabular}
\caption{Master comparison of diffusion, score-based, probability-flow, and flow-matching methods under the unified transport viewpoint. The most important distinctions are: (i) what field is learned, (ii) how the probability path is specified, and (iii) whether sampling is performed with stochastic or deterministic dynamics.}
\label{tab:comparison_master}
\end{table*}

Viewed this way, diffusion and flow matching are less usefully distinguished by broad labels than by a small set of structural decisions: path choice, field parameterization, sampler type, and numerical regime. Recent variants such as cold diffusion, stochastic interpolants, Bayesian Flow Networks, consistency models, and discrete diffusion alternatives suggest that these structural axes extend beyond the original DDPM versus score-SDE versus flow-matching taxonomy, rather than replacing it \citep{bansal2022cold,albergo2025stochastic,graves2023bfn,song2023consistency,kim2024ctm,lou2024sedd}.

\subsection{Design axes and tradeoffs}\label{subsec:comparison_axes}

The methods above can be understood as different choices along a small set of design axes.

\paragraph{(1) What is learned: score versus velocity.}
Diffusion and score-SDE methods learn a score field $s_\theta(x,t)$ and define a sampler via reverse-time dynamics (SDE) or by inducing a velocity through the probability-flow ODE (Section~\ref{sec:pfode}). Flow matching and rectified flow learn a velocity field $v_\theta(x,t)$ directly along a chosen path (Section~\ref{sec:flowmatching}). Score parameterizations are natural for time reversal of diffusions, whereas velocity parameterizations are natural for deterministic transport and can offer more direct numerical control. More recent path-based frameworks such as stochastic interpolants and Bayesian Flow Networks further reinforce this distinction by making the field parameterization itself an explicit part of the modeling design space \citep{albergo2025stochastic,graves2023bfn,xue2024unifyingbfn}.

\paragraph{(2) What path is assumed.}
Diffusion models tie the path $(\rho_t)$ to the marginals of a forward noising process, whereas flow matching makes the path an explicit modeling choice through the coupling/interpolation mechanism. Path choice influences what regions of space and time the model must fit accurately and interacts with time-weighting in the loss (Section~\ref{sec:objectives}). Path design is therefore a primary knob for trading off sample quality, stability, and speed \citep{karras2022edm,lipman2022flowmatching,liu2022rectified}. The same point is illustrated by recent alternatives that change the path itself rather than merely the sampler, including cold diffusion and stochastic interpolants \citep{bansal2022cold,albergo2025stochastic}. Related physics-inspired variants such as PFGM++ reinforce the broader point that path design can be treated as a modeling degree of freedom rather than a fixed consequence of Gaussian noising \citep{xu2023pfgmpp}.

\paragraph{(3) What dynamics are solved at sampling time: SDE versus ODE.}
Reverse-time SDE sampling injects noise at each step, which can aid exploration and sometimes improve perceptual quality, but it also introduces variance and additional discretization considerations. ODE sampling is deterministic given the initial noise draw and can use adaptive solvers, but may be sensitive to stiffness and solver tolerances. Both rely on accurate learned fields and can fail under distribution shift or poor time-weighting \citep{song2021sde}. Newer fast-generation methods such as consistency models and consistency trajectory models can be interpreted as attempts to make this axis less costly by learning transports that remain accurate under very coarse discretization \citep{song2023consistency,kim2024ctm}.

\paragraph{(4) Where error comes from.}
Across methods, generation error can be decomposed into:
(i) \emph{approximation error} in representing the true field (score or velocity),
(ii) \emph{estimation error} from finite data and optimization,
(iii) \emph{numerical error} from discretizing SDEs/ODEs, and
(iv) \emph{path mismatch} when training and sampling use inconsistent assumptions (e.g., different schedules or couplings).
We develop these decompositions more systematically in Section~9. Recent generator-matching and unification perspectives suggest that these error sources may be analyzable within a single broad framework encompassing diffusion, flow matching, and several of their newer variants \citep{patel2024generatormatching,xue2024unifyingbfn}.

\subsection{Practical guidance (qualitative)}\label{subsec:comparison_guidance}

The comparison above does not imply a universal ranking, but it does suggest a few qualitative rules of thumb:
\begin{itemize}
\item If you need a stochastic sampler (diversity, exploration, certain conditioning schemes), reverse-time SDE sampling provides a principled route tied to time reversal \citep{song2021sde}.
\item If you want deterministic sampling and a direct connection to CNF-style likelihood tools, probability-flow ODE sampling provides a natural bridge \citep{song2021sde,chen2018neuralode}.
\item If you want to treat the probability path as a design choice, and potentially reduce stiffness for fast sampling, flow matching and rectified flow offer a direct velocity-learning approach \citep{lipman2022flowmatching,liu2022rectified,lipman2024fmguide}.
\item If your application departs substantially from the standard Gaussian-noising setting---for example, through non-Gaussian degradations, discrete state spaces, or explicitly iterative uncertainty updates---then newer alternatives such as cold diffusion, discrete diffusion, stochastic interpolants, and Bayesian Flow Networks may be better viewed as neighboring points in the same transport design space rather than as entirely separate paradigms \citep{bansal2022cold,lou2024sedd,albergo2025stochastic,graves2023bfn}.
\end{itemize}
These statements are deliberately high-level, and later sections and appendices detail when they hold and where they can fail.

%% file: sections/section9_theory.tex
A unified view of diffusion/score models and flow matching as \emph{learned transport} suggests a common set of theoretical questions. What statistical object does the training objective estimate, and under what weighting? How do approximation and estimation errors in the learned field propagate through reverse-time dynamics? How does numerical discretization bias interact with model error? This section organizes these questions into an error decomposition that is useful for both analysis and practice. We emphasize qualitative structure rather than exhaustive formal results, while pointing to representative recent technical results where appropriate \citep{oko2023minimax,chen2023scoreapprox,zhang2024minimax}.

\subsection{A generic ``field + solver'' abstraction}\label{subsec:theory_field_solver}

Both diffusion/score models and flow matching ultimately generate samples by integrating dynamics driven by a learned field:
\begin{itemize}
\item \textbf{SDE sampler:} $dX_t = b_\theta(X_t,t)\,dt + \sigma(t)\,dW_t$, where $b_\theta$ depends on a learned score (Section~\ref{sec:reverse}).
\item \textbf{ODE sampler:} $\frac{dX_t}{dt} = v_\theta(X_t,t)$, where $v_\theta$ is either induced by a score (probability-flow ODE, Section~\ref{sec:pfode}) or learned directly (flow matching, Section~\ref{sec:flowmatching}).
\end{itemize}
This abstraction separates the overall generative-modeling problem into two coupled components: a \emph{statistical problem}, namely learning an accurate field, and a \emph{numerical problem}, namely integrating the resulting dynamics accurately.

\subsection{Error decomposition: approximation, estimation, and numerics}\label{subsec:theory_error_decomp}

Let $\mu_\theta$ denote the model's implicit sample law obtained by integrating the learned dynamics, exactly or numerically, and let $\mu_0$ denote the target data law. A useful conceptual decomposition of the gap between $\mu_\theta$ and $\mu_0$ is:
\begin{enumerate}
\item \textbf{Approximation error.} Even with infinite data and perfect optimization, the function class for $s_\theta$ or $v_\theta$ may not contain the true field $s_t$ or $v_t$ along the path. This includes limitations due to network architecture, conditioning, and regularity assumptions.
\item \textbf{Estimation/optimization error.} With finite data and stochastic optimization, the learned field deviates from the population minimizer of the training objective (Section~\ref{sec:objectives}). Time weighting can amplify this error in poorly represented SNR regimes.
\item \textbf{Discretization (numerical) error.} Sampling requires discretizing SDE or ODE dynamics. Even with an exact field, finite step sizes introduce bias; with an approximate field, solver error can interact nonlinearly with model error (Sections~\ref{sec:reverse} and \ref{sec:pfode}).
\item \textbf{Path/objective mismatch.} Differences between the path implied by training (noise schedule, coupling, interpolation) and the dynamics used at sampling time (solver choice, step-size schedule, stochastic versus deterministic sampling) can introduce additional error (Sections~\ref{sec:forward} and \ref{sec:comparison}).
\end{enumerate}
This decomposition should be understood as conceptual rather than strictly additive: in practice, these errors interact. Nevertheless, it is useful because it isolates the main mechanisms by which generative quality degrades in finite-compute regimes. Recent statistical analyses of diffusion models have begun to make parts of this picture precise, for example by establishing minimax or near-minimax guarantees under suitable assumptions \citep{oko2023minimax,zhang2024minimax}.

\subsection{Propagation of field error through dynamics}\label{subsec:theory_propagation}

A core theoretical challenge is that small local errors in the learned field can accumulate along trajectories. For ODE sampling, standard stability theory suggests bounds in terms of Lipschitz constants or related conditioning quantities: if $v_\theta$ is close to the true velocity in a suitable norm and the dynamics are well-conditioned, then trajectory error can be controlled. For SDE sampling, one may instead study weak and strong convergence of discretizations together with perturbation bounds for diffusion processes. In either case, the relevant constants can be large in high dimensions or near endpoint regions where the field changes rapidly, which aligns with empirical observations of stiffness and sensitivity to schedules \citep{song2021sde,karras2022edm}.

Recent theory has begun to sharpen this picture for score-based models. For example, score approximation, estimation, and distribution recovery have been analyzed on low-dimensional data and manifold-like settings \citep{chen2023scoreapprox,tang2024adaptivity}, while related work suggests that diffusion models encode nontrivial geometric information about the intrinsic dimension of the underlying data manifold \citep{stanczuk2024intrinsic}. These results remain far from a complete practical theory, but they indicate that field-error propagation can support meaningful geometric and statistical guarantees.

\subsection{Discretization bias and solver choice}\label{subsec:theory_discretization}

Sampling error is often dominated by discretization. Euler--Maruyama introduces $O(\Delta t)$ weak error for SDEs under regularity assumptions, while higher-order methods can reduce this at increased computational cost. For ODEs, higher-order Runge--Kutta solvers or adaptive solvers can reduce local truncation error, but may still struggle with stiffness (Section~\ref{subsec:pfode_sampling}). In practice, the cost--quality tradeoff is governed by (i) how rapidly the learned field varies across time and space and (ii) how strongly error concentrates near endpoint regions. This is one reason few-step generation is difficult: it amounts to controlling numerical error under deliberately coarse discretization. These considerations motivate work on better schedules and better path design, as in EDM and rectified flow \citep{karras2022edm,liu2022rectified}.

More recent one-step and few-step models make this issue especially explicit. Consistency models, one-step score distillation, and EM-style distillation can all be viewed as attempts to reshape the learning problem so that numerically coarse samplers remain accurate \citep{song2023consistency,luo2024sim,xie2024emd}. This suggests that fast sampling is not merely an implementation concern, but a core theoretical question about how field learning and discretization interact.

\subsection{Identifiability and what the objective recovers}\label{subsec:theory_identifiability}

An often underemphasized point is that training objectives recover a field \emph{under the training distribution}. For diffusion and score-based models, DSM trains the score of perturbed marginals $\rho_t$, not the score of $\rho_0$ directly, and success depends on accurately learning $s_t$ over the time region that most influences reverse-time sampling (Section~\ref{sec:objectives}). For flow matching, the learned velocity is tied to the chosen coupling and path sampler (Section~\ref{sec:flowmatching}), and changing the coupling can change the learned velocity even when the endpoint laws are fixed. Thus identifiability is path-dependent: there is typically no unique ``best'' field without first fixing the path measure.

This observation becomes even more important in more recent variants. For example, one-step distillation, Gaussian-mixture flow matching, and manifold-aware diffusion analyses all reinforce the idea that identifiability depends not only on the endpoint law, but also on the chosen state representation, path geometry, and objective weighting \citep{chen2025gmflow,stanczuk2024intrinsic}.

\subsection{What theory is still missing (and why it is hard)}\label{subsec:theory_missing}

Despite rapid empirical progress, several fundamental questions remain open or only partially addressed:
\begin{itemize}
\item \textbf{High-dimensional guarantees.} How do approximation and sampling complexity scale with dimension for realistic, low-dimensional data manifolds embedded in high-dimensional spaces \citep{oko2023minimax,zhang2024minimax}?
\item \textbf{Generalization of learned fields.} When does a learned score or velocity generalize off the training support, and how does this affect sampling stability?
\item \textbf{Coupled model--solver analysis.} Most existing analyses isolate either learning error (assuming exact sampling) or numerical error (assuming exact fields). Practical systems require joint bounds.
\item \textbf{Principled path design.} Which path choices optimize stability and sample quality under compute constraints? Rectified flow, Gaussian-mixture flow matching, and diffusion-design work all suggest the importance of this question, but a unifying theory is still developing \citep{karras2022edm,liu2022rectified,chen2025gmflow}.
\item \textbf{Geometry-aware theory.} To what extent should diffusion and flow-matching theory be stated relative to ambient Euclidean dimension, and to what extent can it adapt to intrinsic manifold structure \citep{tang2024adaptivity,stanczuk2024intrinsic}?
\end{itemize}

We view these gaps as opportunities rather than merely limitations. Once diffusion, score-based models, and flow matching are all understood as instances of learned transport, they can be analyzed using a common language of fields, paths, and solvers. In that sense, theoretical progress on any one of these paradigms is likely to inform the others as well.

%% file: sections/section10_open_problems.tex
We close by summarizing research directions suggested by the unified transport viewpoint developed throughout this survey. These problems are intentionally phrased in a method-agnostic way: they apply across diffusion and score-based models, probability-flow ODE samplers, flow matching, rectified flows, and a broader family of discrete-state, graph-based, and non-Euclidean generative transports \citep{campbell2022ctddm,chen2023analogbits,gat2024dfm,kerrigan2024ffm,haviv2025wfm}. They also interact strongly with the question of \emph{conditioning}, since constrained generation, inverse problems, editing, and restoration all modify transport dynamics in ways that are sensitive to both model geometry and numerical implementation \citep{meng2022sdedit,lugmayr2022repaint,tewari2023forwardmodels}.

\subsection{Principled path and schedule design}\label{subsec:open_path_design}

Path choice---whether implemented as a diffusion noise schedule or as an explicit coupling/interpolation in flow matching---appears to be a first-order determinant of numerical stability and sampling efficiency. A major open problem is to develop \emph{principled} criteria for path design under compute constraints. Examples include choosing paths that reduce stiffness, concentrate modeling capacity where it matters most for generation, or yield robust conditioning behavior for inverse problems. Recent empirical and conceptual proposals, including EDM, rectified flow, critically damped Langevin diffusion, cold diffusion, Poisson-flow-style transports, and alternative discrete-state noising schemes, highlight the importance of this degree of freedom, but a comprehensive theory remains incomplete \citep{karras2022edm,liu2022rectified,dockhorn2022cld,bansal2022cold,xu2022pfgm,austin2021d3pm,okhotin2023ssddpm}.

\subsection{Coupled learning--sampling analysis}\label{subsec:open_coupled_analysis}

Practical generative modeling combines an estimated field (score or velocity) with an approximate numerical solver. Developing bounds that jointly capture \emph{learning error} and \emph{solver error}, together with their interaction, is essential for explaining when few-step sampling succeeds and when it fails. This issue becomes even more pressing as newer methods explicitly target coarse discretization, direct trajectory learning, or alternative forward processes. A broader theoretical challenge is therefore to understand diffusion, flow matching, autoregressive diffusion, and related generator families within a common field-plus-solver framework, rather than analyzing each architecture in isolation \citep{hoogeboom2022ardm,dockhorn2022cld,liu2022rectified,lipman2022flowmatching}.

\subsection{Generalization and robustness of learned fields}\label{subsec:open_generalization}

Because training objectives estimate fields under a particular path distribution (Section~\ref{subsec:theory_identifiability}), it remains unclear when learned scores or velocities generalize to off-support regions encountered during sampling, guidance, or conditioning. Understanding the geometry of learned vector fields, their regularity, and their behavior under perturbations could yield both better theoretical guarantees and practical diagnostics for robust generation. This issue is likely to become even sharper as the state spaces under consideration broaden, for example to discrete domains, graph spaces, function spaces, or spaces of probability distributions \citep{campbell2022ctddm,xu2024discretectgraph,kerrigan2024ffm,haviv2025wfm}.

\subsection{Fast sampling and distillation beyond heuristics}\label{subsec:open_fast_sampling}

Fast generation remains a central practical objective. Current progress relies heavily on improved discretizations, solver choices, path design, and distillation-like procedures. A promising direction is to formalize fast sampling itself as a constrained transport problem: given a fixed compute budget---for example, a limit on the number of function evaluations---choose a path and field parameterization that minimizes distributional error. Recent one-step diffusion methods, score-implicit distillation, and EM-style distillation all suggest that this problem can be attacked at the level of objective design rather than only at the level of numerical implementation \citep{song2023consistency,luo2024sim,xie2024emd}. A unified theory of this tradeoff is still missing.

\subsection{Conditioning, inverse problems, and constrained transport}\label{subsec:open_conditioning}

For inverse problems and conditional generation, conditioning can be interpreted as modifying the transport dynamics by incorporating a likelihood or constraint term along the path (Section~\ref{sec:intro}). The transport lens therefore suggests viewing these methods as instances of \emph{constrained} or \emph{controlled} transport. This raises natural questions about stability, bias, and optimality of conditioning schemes under different sampler choices (SDE versus ODE) and path designs. Schr\"odinger bridge formulations provide one principled route by optimizing over stochastic path measures under endpoint constraints \citep{debortoli2021dsb}. More recent work on guided image editing, inpainting, restoration, forward-model-based inverse problems, and diffusion posterior sampling suggests that this area is likely to become an important point of convergence between diffusion, bridge, and flow-matching approaches \citep{meng2022sdedit,lugmayr2022repaint,tewari2023forwardmodels,rout2023latentinverse,zhang2023ucdir,chung2023dps}.

A particularly promising direction is to understand inverse problems through the joint lens of posterior sampling, latent diffusion priors, expectation-maximization, and diffusion optimal control \citep{rout2023latentinverse,rozet2024emprior,li2024diffoptcontrol}. Related work also suggests that the computational bottleneck in inverse problems is often not only learning the prior, but performing posterior transport efficiently under strong data constraints \citep{janati2024dcps,pandey2024fastinverse}. This points toward a richer theory of conditioned transport in which path design, solver choice, and constraint handling are treated jointly rather than separately.

\subsection{Beyond Gaussian, Euclidean, and continuous-state formulations}\label{subsec:open_beyond_standard}

A major emerging direction is to understand how far the transport viewpoint can be pushed beyond the standard setting of continuous Gaussian noising in Euclidean data spaces. Recent work has explored autoregressive diffusion \citep{hoogeboom2022ardm}, continuous-time denoising models for discrete data \citep{campbell2022ctddm}, structured discrete-state diffusion \citep{austin2021d3pm}, discrete diffusion via continuous embeddings and masking mechanisms \citep{chen2023analogbits,sahoo2024simplemaskedlm}, graph-oriented discrete-state transports \citep{xu2024discretectgraph,eijkelboom2024vfm}, and flow matching in discrete, functional, and Wasserstein spaces \citep{gat2024dfm,cheng2024sfm,davis2024fisherflow,kerrigan2024ffm,haviv2025wfm}. These developments suggest that the basic design axes emphasized in this survey---path, learned field, and sampler---survive far beyond the original DDPM setting. A major open problem is to identify which parts of the existing theory and practice genuinely generalize and which depend crucially on Gaussian or Euclidean structure.

\subsection{Evaluation: what should we measure?}\label{subsec:open_evaluation}

Finally, evaluation remains a persistent challenge. In practice, generative models are often judged by metrics such as the Fr\'echet Inception Distance (FID) \citep{heusel2017fid}, precision--recall style metrics \citep{sajjadi2018precision,kynkaanniemi2019improvedpr,simon2019revisitingpr,cheema2023prc,liang2024efficientpr}, and related fidelity/diversity decompositions \citep{naeem2020reliable}. While these metrics are useful, they often conflate distinct notions of quality, including sample fidelity, coverage, perceptual realism, and support overlap. More recent work has therefore proposed sample-level, likelihood-like, or attribute-based alternatives that attempt to evaluate generalization, interpretability, or faithfulness more directly \citep{alaa2022faithful,jiralerspong2023fld,kim2024attributeeval}.

From the perspective of this survey, the central problem is that standard endpoint metrics may be insensitive to failure modes that matter for downstream tasks such as inverse problems, controlled generation, or scientific applications. In addition, several recent papers argue that widely used metrics can behave asymmetrically or unfairly, especially in high-dimensional settings or when comparing different classes of samplers such as diffusion and flow-based models \citep{khayatkhoei2023asymmetry,stein2023flaws,raisa2025positionmetrics}. A transport-based perspective therefore suggests evaluation criteria that reflect not only endpoint distributional accuracy, but also the stability, regularity, controllability, and solver sensitivity of the learned dynamics. This challenge becomes even sharper as the design space broadens to include autoregressive diffusion, non-Gaussian degradations, discrete-state diffusion, non-Euclidean transport formulations, and conditioning-intensive inverse-problem solvers \citep{hoogeboom2022ardm,bansal2022cold,campbell2022ctddm,chen2023analogbits,sahoo2024simplemaskedlm,kerrigan2024ffm,haviv2025wfm,pandey2024fastinverse}.

\paragraph{Takeaway.}
Diffusion and score-based models, probability-flow ODEs, and flow matching can all be understood as complementary parameterizations of probability transport. The open problems above---especially those concerning path design, coupled learning--sampling theory, robust conditioning, and extensions beyond Gaussian Euclidean settings---are therefore not isolated technical questions, but shared challenges across multiple generative-modeling paradigms. Progress on these questions is likely to clarify which aspects of performance are intrinsic to the learned field and which are artifacts of the chosen path and numerical solver.

%% file: sections/appendixA_reverse_time_diffusions.tex
This appendix provides background for Proposition~\ref{prop:reverse_sde}. We state the reverse-time diffusion formula at a formal level and explain why the score $\nabla_x \log \rho_t(x)$ appears naturally in the reverse drift.

\subsection{Setup and assumptions}

Consider the forward It\^{o} diffusion
\begin{equation}\label{eq:app_forward_sde}
dX_t = f(X_t,t)\,dt + g(t)\,dW_t, \qquad t\in[0,1],
\end{equation}
where $f:\mathbb{R}^d\times[0,1]\to\mathbb{R}^d$ is a drift field, $g:[0,1]\to\mathbb{R}_+$ is a scalar diffusion coefficient, and $(W_t)_{t\in[0,1]}$ is a standard Wiener process. Let $\mu_t$ denote the law of $X_t$, and assume that for each $t\in(0,1]$, $\mu_t$ admits a sufficiently smooth positive density $\rho_t$.

We do not attempt to state minimal assumptions here. Informally, one needs enough regularity to justify both the Fokker--Planck equation and the time-reversal argument, including existence of smooth transition densities and suitable decay or boundary behavior. Classical references include \citet{anderson1982reverse}; the formulation used in modern score-based modeling is presented by \citet{song2021sde}.

\subsection{Forward density evolution}

When densities exist, the forward marginals satisfy the Fokker--Planck equation
\begin{equation}\label{eq:app_fp}
\partial_t \rho_t(x)
=
-\nabla\cdot\big(\rho_t(x)f(x,t)\big)
+
\tfrac12 g(t)^2 \Delta \rho_t(x).
\end{equation}
It is useful to rewrite the diffusion term using
\[
\Delta \rho_t
=
\nabla\cdot\!\big(\rho_t \nabla \log \rho_t\big),
\]
which gives
\begin{equation}\label{eq:app_fp_rewrite}
\partial_t \rho_t
=
-\nabla\cdot\!\Big(\rho_t f - \tfrac12 g(t)^2 \rho_t \nabla \log \rho_t\Big)
-\tfrac12 g(t)^2 \nabla\cdot\!\big(\rho_t \nabla \log \rho_t\big)
+
\tfrac12 g(t)^2 \Delta \rho_t.
\end{equation}
Formally, this makes visible the velocity-like correction involving the score.

\subsection{Reverse-time dynamics}

Let $\widetilde X_t := X_{1-t}$ denote the time-reversed process. Under suitable regularity assumptions, the reverse-time process is again a diffusion. Its drift is not simply the negative of the forward drift; instead, it contains an additional correction term involving the score of the forward marginals. Formally, the reverse-time SDE can be written as
\begin{equation}\label{eq:app_reverse_sde}
dX_t
=
\Big(f(X_t,t)-g(t)^2 \nabla_x \log \rho_t(X_t)\Big)\,dt
+
g(t)\,d\bar W_t,
\end{equation}
interpreted in reverse time, where $\bar W_t$ is a reverse-time Wiener process.

Equation~\eqref{eq:app_reverse_sde} is the key fact underlying score-based generative modeling. It shows that reverse-time sampling can be defined once the score field
\[
s_t(x)=\nabla_x \log \rho_t(x)
\]
is known or accurately approximated.

\subsection{Why the score appears}

A formal way to understand the appearance of the score is to compare the forward Fokker--Planck equation with the continuity equation that would govern reversed mass transport. The diffusion term in \eqref{eq:app_fp} cannot be reversed simply by negating time; the stochastic spreading of mass must be compensated by an additional drift that points toward regions of higher probability. That compensating drift is precisely the score term.

Equivalently, one may think of the reverse drift as the forward drift corrected by the logarithmic gradient of the evolving density:
\[
\text{reverse drift}
=
f(x,t)-g(t)^2 \nabla_x \log \rho_t(x).
\]
The factor $g(t)^2$ reflects the strength of the diffusion in the forward process. Larger forward noise requires a stronger score correction in reverse time.

\subsection{Connection to score-based generative modeling}

Modern score-based models replace the unknown score $s_t(x)=\nabla_x \log \rho_t(x)$ with a neural approximation $s_\theta(x,t)$. Substituting $s_\theta$ into \eqref{eq:app_reverse_sde} gives the practical reverse-time sampler used in score-SDE modeling:
\[
dX_t
=
\Big(f(X_t,t)-g(t)^2 s_\theta(X_t,t)\Big)\,dt
+
g(t)\,d\bar W_t.
\]
This is the mathematical reason score estimation suffices for generation: one need not represent the density $\rho_t$ itself, only its logarithmic gradient along the forward path.

\subsection{Remark on conventions}

Different references write the reverse-time diffusion with different sign conventions, depending on whether time is parameterized forward or backward and how the reversed Brownian motion is defined. These formulations are equivalent after a change of variables, but care is needed when comparing formulas across sources. Throughout this paper, we follow the convention of \citet{song2021sde}, in which the reverse-time drift is written at time $t$ and sampling proceeds from $t=1$ to $t=0$.

%% file: sections/appendixB_pfode_and_pdes.tex
This appendix provides background for Proposition~\ref{prop:pfode}. The goal is to make explicit the PDE relationship between stochastic diffusion dynamics, deterministic transport, and the probability-flow ODE. The key point is that the probability-flow ODE is constructed so that its one-time marginals satisfy the same density evolution equation as the forward SDE, even though the induced path measures generally differ.

\subsection{Forward diffusion and the Fokker--Planck equation}

Consider the forward It\^{o} diffusion
\begin{equation}\label{eq:appB_forward_sde}
dX_t = f(X_t,t)\,dt + g(t)\,dW_t,
\end{equation}
where $f:\mathbb{R}^d\times[0,1]\to\mathbb{R}^d$ is a drift field, $g:[0,1]\to\mathbb{R}_+$ is a scalar diffusion coefficient, and $(W_t)_{t\in[0,1]}$ is a standard Wiener process. Let $\mu_t$ denote the law of $X_t$, and assume that $\mu_t$ admits a sufficiently smooth density $\rho_t$.

Under standard regularity assumptions, $\rho_t$ satisfies the Fokker--Planck equation
\begin{equation}\label{eq:appB_fp}
\partial_t \rho_t(x)
=
-\nabla\cdot\!\big(\rho_t(x)f(x,t)\big)
+
\tfrac12 g(t)^2 \Delta \rho_t(x).
\end{equation}
This equation describes the evolution of the one-time marginals of the stochastic process \eqref{eq:appB_forward_sde}.

\subsection{Deterministic transport and the continuity equation}

Now consider a deterministic flow defined by the ODE
\begin{equation}\label{eq:appB_ode}
\frac{dX_t}{dt} = v(X_t,t),
\end{equation}
where $v:\mathbb{R}^d\times[0,1]\to\mathbb{R}^d$ is a velocity field. If $X_0\sim\mu_0$ and the flow is well-defined, then the corresponding densities evolve according to the continuity equation
\begin{equation}\label{eq:appB_continuity}
\partial_t \rho_t(x) + \nabla\cdot\!\big(\rho_t(x)v(x,t)\big)=0.
\end{equation}
Equivalently,
\begin{equation}\label{eq:appB_continuity_rewrite}
\partial_t \rho_t(x)
=
-\nabla\cdot\!\big(\rho_t(x)v(x,t)\big).
\end{equation}
The central question is therefore: can one choose a deterministic velocity field $v$ so that \eqref{eq:appB_continuity_rewrite} matches the Fokker--Planck evolution \eqref{eq:appB_fp}?

\subsection{Rewriting the diffusion term}

The answer rests on the identity
\begin{equation}\label{eq:appB_laplacian_identity}
\Delta \rho_t
=
\nabla\cdot\!\big(\rho_t \nabla \log \rho_t\big),
\end{equation}
valid whenever $\rho_t$ is positive and sufficiently smooth. Substituting \eqref{eq:appB_laplacian_identity} into the Fokker--Planck equation gives
\begin{equation}\label{eq:appB_fp_rewrite}
\partial_t \rho_t
=
-\nabla\cdot\!\big(\rho_t f\big)
+
\tfrac12 g(t)^2 \nabla\cdot\!\big(\rho_t \nabla \log \rho_t\big).
\end{equation}
Factoring out the divergence yields
\begin{equation}\label{eq:appB_fp_flux}
\partial_t \rho_t
=
-\nabla\cdot\!\left(
\rho_t f
-
\tfrac12 g(t)^2 \rho_t \nabla \log \rho_t
\right).
\end{equation}
Comparing \eqref{eq:appB_fp_flux} with the continuity equation \eqref{eq:appB_continuity_rewrite}, one is led to define the deterministic velocity field
\begin{equation}\label{eq:appB_pf_velocity}
v_{\mathrm{PF}}(x,t)
=
f(x,t)-\tfrac12 g(t)^2 \nabla_x \log \rho_t(x).
\end{equation}

\subsection{Probability-flow ODE}

The corresponding deterministic dynamics are
\begin{equation}\label{eq:appB_pfode}
\frac{dX_t}{dt}
=
f(X_t,t)-\tfrac12 g(t)^2 \nabla_x \log \rho_t(X_t).
\end{equation}
By construction, the continuity equation associated with \eqref{eq:appB_pfode} is exactly \eqref{eq:appB_fp_flux}, which is the same density evolution equation satisfied by the forward SDE. Therefore the ODE \eqref{eq:appB_pfode} and the SDE \eqref{eq:appB_forward_sde} share the same one-time marginals.

This proves Proposition~\ref{prop:pfode} at a formal level: the probability-flow ODE is defined precisely so that its density evolution matches the Fokker--Planck evolution of the SDE.

\subsection{Marginals versus path measures}

It is important to emphasize what Proposition~\ref{prop:pfode} does and does not say. The proposition states that the SDE and ODE have the same one-time marginals:
\[
\mu_t^{\mathrm{SDE}} = \mu_t^{\mathrm{ODE}}
\qquad \text{for each } t\in[0,1].
\]
It does \emph{not} say that they induce the same law on trajectories. The SDE defines a stochastic path measure, while the ODE defines a deterministic flow map. Thus the two models agree at the level of marginals but generally differ at the level of paths. This is the distinction emphasized in Section~\ref{subsec:marginals_vs_paths}.

\subsection{Learned probability-flow ODE}

In practice, the score $\nabla_x\log \rho_t(x)$ is unknown and is replaced by a learned approximation $s_\theta(x,t)$. This yields the practical probability-flow ODE
\begin{equation}\label{eq:appB_pfode_theta}
\frac{dX_t}{dt}
=
f(X_t,t)-\tfrac12 g(t)^2 s_\theta(X_t,t).
\end{equation}
When $s_\theta\approx \nabla_x\log \rho_t$, the dynamics \eqref{eq:appB_pfode_theta} approximately preserve the intended one-time marginals. The quality of the resulting sampler then depends on both score approximation error and numerical integration error.

\subsection{Instantaneous change-of-variables}

Suppose that the learned velocity field
\[
v_\theta(x,t)=f(x,t)-\tfrac12 g(t)^2 s_\theta(x,t)
\]
defines an invertible flow. Then the corresponding density evolves according to the instantaneous change-of-variables formula
\begin{equation}\label{eq:appB_icov}
\frac{d}{dt}\log \rho_t(X_t)
=
-\nabla\cdot v_\theta(X_t,t).
\end{equation}
This follows directly from the continuity equation and is the continuous-time analogue of the standard change-of-variables rule used in normalizing flows \citep{chen2018neuralode}. Integrating \eqref{eq:appB_icov} along a trajectory yields
\begin{equation}\label{eq:appB_loglikelihood}
\log \rho_0(X_0)
=
\log \rho_1(X_1)
+
\int_0^1 \nabla\cdot v_\theta(X_t,t)\,dt,
\end{equation}
up to the direction-of-time convention used to parameterize the flow. This is the basis for the likelihood connection discussed in Section~\ref{sec:pfode}.

\subsection{Remark on regularity}

The derivations above are formal and assume sufficient smoothness, positivity of densities, and existence of the required derivatives. In particular, the identity \eqref{eq:appB_laplacian_identity} and the change-of-variables formula \eqref{eq:appB_icov} both require regularity conditions that may fail in singular or degenerate settings. For the purposes of this survey, the main point is conceptual: the probability-flow ODE is obtained by rewriting the Fokker--Planck equation as a continuity equation with a score-corrected deterministic velocity.

\subsection{Takeaway}

The probability-flow ODE does not represent a different endpoint problem from the forward SDE. Rather, it is a deterministic reparameterization of the same marginal evolution. This is why it provides a natural bridge between diffusion models and deterministic transport methods such as continuous normalizing flows: the score field learned for stochastic reverse-time sampling can also be used to define a deterministic ODE sampler with the same one-time marginals.

%% file: sections/appendixC_objective_equivalences.tex
This appendix provides background for Proposition~\ref{prop:weighted_fisher}. The goal is to make explicit why denoising score matching (DSM), DDPM-style noise-prediction losses, and continuous-time score-SDE objectives can all be viewed as weighted score-regression objectives under Gaussian perturbations. We keep the derivations at a survey level, emphasizing the common structure rather than the most general technical formulation \citep{hyvarinen2005score,vincent2011dsm,ho2020ddpm,song2021sde,kingma2021vdm}.

\subsection{Classical score matching and Fisher divergence}
\label{subsec:appC_score_matching}

Let $\rho$ be a target density on $\mathbb{R}^d$ and let $s_\theta:\mathbb{R}^d\to\mathbb{R}^d$ be a model score field. The ideal Fisher-divergence objective is
\begin{equation}\label{eq:appC_fisher}
\mathcal{J}_{\mathrm{F}}(\theta)
=
\frac12\,\mathbb{E}_{x\sim \rho}\!\left[\|s_\theta(x)-\nabla_x \log \rho(x)\|^2\right].
\end{equation}
At first sight, \eqref{eq:appC_fisher} appears intractable because the true score $\nabla_x \log \rho(x)$ is unknown. The key observation of \citet{hyvarinen2005score} is that, under suitable regularity and boundary conditions, \eqref{eq:appC_fisher} can be rewritten by integration by parts into a form that depends only on samples from $\rho$ and the model score:
\begin{equation}\label{eq:appC_hyvarinen}
\mathcal{J}_{\mathrm{F}}(\theta)
\equiv
\mathbb{E}_{x\sim \rho}\!\left[
\nabla\cdot s_\theta(x)
+
\frac12 \|s_\theta(x)\|^2
\right]
+
\text{constant},
\end{equation}
where the constant is independent of $\theta$.

Thus score matching estimates the logarithmic gradient of a density without ever requiring direct evaluation of the density itself. In modern generative modeling, this basic idea is not applied to a single density $\rho$, but to a family of perturbed marginals $(\rho_t)_{t\in[0,1]})$.

\subsection{Denoising score matching under a corruption kernel}
\label{subsec:appC_dsm}

Let $q_t(x_t\mid x_0)$ be a tractable corruption kernel and let $\rho_0$ denote the data density. The corresponding perturbed marginal is
\begin{equation}\label{eq:appC_marginal}
\rho_t(x_t)=\int q_t(x_t\mid x_0)\,\rho_0(x_0)\,dx_0.
\end{equation}
A key identity underlying denoising score matching is
\begin{equation}\label{eq:appC_dsm_identity}
\nabla_{x_t}\log \rho_t(x_t)
=
\mathbb{E}\!\left[\nabla_{x_t}\log q_t(x_t\mid x_0)\,\middle|\,x_t\right].
\end{equation}
This identity shows that the score of the perturbed marginal can be recovered as a conditional expectation of the conditional score. Therefore one may train a model score field by regressing against the analytically available conditional target rather than the intractable marginal score \citep{vincent2011dsm}.

The denoising score-matching objective is
\begin{equation}\label{eq:appC_dsm_loss}
\mathcal{L}_{\mathrm{DSM}}(\theta)
=
\mathbb{E}_{x_0\sim \rho_0}\,
\mathbb{E}_{x_t\sim q_t(\cdot\mid x_0)}
\Big[
\|s_\theta(x_t,t)-\nabla_{x_t}\log q_t(x_t\mid x_0)\|^2
\Big].
\end{equation}
By the $L^2$ projection identity,
\begin{align}
\mathcal{L}_{\mathrm{DSM}}(\theta)
&=
\mathbb{E}\!\left[
\big\|s_\theta(x_t,t)-\mathbb{E}[\nabla_{x_t}\log q_t(x_t\mid x_0)\mid x_t]\big\|^2
\right]
+
\text{constant} \notag\\
&=
\mathbb{E}\!\left[
\|s_\theta(x_t,t)-\nabla_{x_t}\log \rho_t(x_t)\|^2
\right]
+
\text{constant}.
\label{eq:appC_dsm_projection}
\end{align}
Hence the population minimizer of DSM is the score of the perturbed marginal:
\[
s_\theta^*(x_t,t)=\nabla_{x_t}\log \rho_t(x_t).
\]

This is the key sense in which DSM is already a score-regression objective on perturbed data.

\subsection{Gaussian perturbations}
\label{subsec:appC_gaussian}

Now assume the perturbation kernel is Gaussian:
\begin{equation}\label{eq:appC_gaussian_kernel}
x_t = m(t)x_0 + s(t)\varepsilon,
\qquad
\varepsilon\sim \mathcal{N}(0,I).
\end{equation}
Then
\[
q_t(x_t\mid x_0)
=
\mathcal{N}\!\big(m(t)x_0,\ s(t)^2 I\big),
\]
and its conditional score is
\begin{equation}\label{eq:appC_conditional_score}
\nabla_{x_t}\log q_t(x_t\mid x_0)
=
-\frac{1}{s(t)^2}\big(x_t-m(t)x_0\big).
\end{equation}
Using \eqref{eq:appC_gaussian_kernel}, we also have
\begin{equation}\label{eq:appC_conditional_score_eps}
x_t-m(t)x_0=s(t)\varepsilon,
\qquad\text{hence}\qquad
\nabla_{x_t}\log q_t(x_t\mid x_0)
=
-\frac{1}{s(t)}\,\varepsilon.
\end{equation}
This is the basic algebraic reason that Gaussian DSM can be written either as score regression or as noise regression.

\subsection{Continuous-time weighted DSM}
\label{subsec:appC_ct_dsm}

In the continuous-time setting, one samples $t$ from a distribution on $[0,1]$ and minimizes a weighted objective of the form
\begin{equation}\label{eq:appC_ct_dsm}
\mathcal{L}_{\mathrm{CT}}(\theta)
=
\mathbb{E}_{t}\,
\mathbb{E}_{x_0,\varepsilon}
\Big[
\lambda(t)\,
\|s_\theta(x_t,t)-\nabla_{x_t}\log q_t(x_t\mid x_0)\|^2
\Big],
\end{equation}
where $x_t$ is generated according to \eqref{eq:appC_gaussian_kernel}. Substituting \eqref{eq:appC_conditional_score} gives
\begin{equation}\label{eq:appC_ct_dsm_expanded}
\mathcal{L}_{\mathrm{CT}}(\theta)
=
\mathbb{E}_{t}\,
\mathbb{E}_{x_0,\varepsilon}
\Big[
\lambda(t)\,
\left\|
s_\theta(x_t,t)
+
\frac{1}{s(t)^2}\big(x_t-m(t)x_0\big)
\right\|^2
\Big],
\end{equation}
or equivalently, using \eqref{eq:appC_conditional_score_eps},
\begin{equation}\label{eq:appC_ct_dsm_eps}
\mathcal{L}_{\mathrm{CT}}(\theta)
=
\mathbb{E}_{t}\,
\mathbb{E}_{x_0,\varepsilon}
\Big[
\lambda(t)\,
\left\|
s_\theta(x_t,t)
+
\frac{1}{s(t)}\varepsilon
\right\|^2
\Big].
\end{equation}

Because the population minimizer is the score of the perturbed marginal $\rho_t$, this is a weighted score-regression objective along the diffusion path \citep{song2021sde}.

\subsection{Weighted Fisher-divergence interpretation}
\label{subsec:appC_weighted_fisher}

The previous subsection implies the weighted Fisher-divergence viewpoint used in Section~\ref{sec:objectives}. Indeed, applying \eqref{eq:appC_dsm_projection} pointwise in time yields
\begin{equation}\label{eq:appC_weighted_fisher}
\mathcal{L}_{\mathrm{CT}}(\theta)
\equiv
\int_0^1
\lambda(t)\,
\mathbb{E}_{x\sim \rho_t}
\big[
\|s_\theta(x,t)-s_t(x)\|^2
\big]
\,dt
+
\text{constant},
\end{equation}
where
\[
s_t(x)=\nabla_x \log \rho_t(x).
\]
Thus the continuous-time objective is, up to an additive constant independent of $\theta$, a time-integrated Fisher divergence between the learned score and the true score of the perturbed marginals.

This is the formal content behind Proposition~\ref{prop:weighted_fisher}: different diffusion and score-based objectives are unified because they estimate the same statistical object, namely the score field of $(\rho_t)$, with different parameterizations and weights.

\subsection{DDPM: ELBO versus simplified noise prediction}
\label{subsec:appC_ddpm}

DDPM is derived from a variational lower bound on the reverse Markov-chain likelihood \citep{sohldickstein2015deep,ho2020ddpm}. In discrete time, the full objective can be written as a sum of KL and reconstruction terms:
\begin{align}
\mathcal{L}_{\mathrm{ELBO}}
&=
\mathbb{E}\Big[
\mathrm{KL}\big(q(x_T\mid x_0)\,\|\,p(x_T)\big)
+
\sum_{t=2}^{T}
\mathrm{KL}\big(q(x_{t-1}\mid x_t,x_0)\,\|\,p_\theta(x_{t-1}\mid x_t)\big)
\notag\\
&\hspace{2.0cm}
-\log p_\theta(x_0\mid x_1)
\Big].
\label{eq:appC_elbo}
\end{align}
In practice, however, the most widely used training loss is the simplified noise-prediction objective
\begin{equation}\label{eq:appC_ddpm_eps_loss}
\mathcal{L}_{\varepsilon}(\theta)
=
\mathbb{E}_{t,x_0,\varepsilon}
\big[
\|\varepsilon-\varepsilon_\theta(x_t,t)\|^2
\big],
\qquad
x_t=\sqrt{\bar\alpha_t}x_0+\sqrt{1-\bar\alpha_t}\varepsilon.
\end{equation}
This practical loss is a reweighted surrogate of the variational objective rather than a term-by-term identical rewriting of \eqref{eq:appC_elbo}. Later work clarified this weighting structure and its continuous-time limit \citep{nichol2021improved,kingma2021vdm}.

\subsection{DDPM noise prediction as score regression}
\label{subsec:appC_ddpm_score}

In DDPM notation, one identifies
\[
m(t)=\sqrt{\bar\alpha_t},
\qquad
s(t)=\sqrt{1-\bar\alpha_t},
\]
so the conditional score becomes
\begin{equation}\label{eq:appC_ddpm_conditional_score}
\nabla_{x_t}\log q_t(x_t\mid x_0)
=
-\frac{1}{\sqrt{1-\bar\alpha_t}}\,\varepsilon.
\end{equation}
Now let $\varepsilon_\theta(x_t,t)$ denote a model that predicts the noise. Define the induced score model by
\begin{equation}\label{eq:appC_eps_to_score}
s_\theta(x_t,t)
=
-\frac{1}{\sqrt{1-\bar\alpha_t}}\,\varepsilon_\theta(x_t,t).
\end{equation}
Then minimizing \eqref{eq:appC_ddpm_eps_loss} is equivalent, up to a schedule-dependent scaling, to minimizing squared error between $s_\theta(x_t,t)$ and the conditional score \eqref{eq:appC_ddpm_conditional_score}. Thus DDPM noise prediction is a reparameterized form of score regression \citep{ho2020ddpm,kingma2021vdm}.

\subsection{Parameterization equivalences}
\label{subsec:appC_param_equiv}

For Gaussian perturbations, one may move freely between three common parameterizations:
\begin{itemize}
\item \textbf{Noise prediction:} predict $\varepsilon$ in
\[
x_t=m(t)x_0+s(t)\varepsilon.
\]
\item \textbf{Clean-sample prediction:} solve for
\[
x_0=\frac{x_t-s(t)\varepsilon}{m(t)}.
\]
\item \textbf{Score prediction:} use
\[
s_t(x_t\mid x_0) = -\frac{1}{s(t)}\varepsilon.
\]
\end{itemize}
These parameterizations are algebraically equivalent in the Gaussian setting, but they lead to different optimization geometries because the effective scaling with respect to $t$ changes. This is one reason alternative parameterizations can show different empirical behavior despite corresponding to the same underlying score field \citep{nichol2021improved,song2020ddim,kingma2021vdm,salimans2022progressive}.

\subsection{Takeaway}
\label{subsec:appC_takeaway}

The main conclusion of this appendix is that many objectives used in diffusion and score-based generative modeling are not fundamentally different estimators of different quantities. Rather, they are different weighted and reparameterized ways of estimating the score field of the perturbed marginals. This is the sense in which DDPM noise-prediction losses, DSM, and continuous-time score-SDE objectives are unified by Proposition~\ref{prop:weighted_fisher}.

%% file: sections/appendixD_flow_matching_derivations.tex
This appendix provides background for Proposition~\ref{prop:fm_velocity}. The goal is to make explicit the relationship between endpoint-conditioned regression targets in flow matching and the marginal velocity field that appears in the continuity equation. The key point is that the practical conditional flow-matching objective is an $L^2$ regression problem whose population minimizer is the conditional expectation of the path derivative given the current state and time \citep{lipman2022flowmatching,lipman2024fmguide,albergo2025stochastic}.

\subsection{Path samplers and induced probability paths}
\label{subsec:appD_path_sampler}

Let $\mu_0$ denote the data law and $\mu_1$ a reference law, typically Gaussian. Let
\[
\pi(x_0,x_1)
\]
be a coupling between $\mu_0$ and $\mu_1$. A path sampler specifies, for each $t\in[0,1]$, a random intermediate state
\[
x_t = \Phi_t(x_0,x_1),
\]
where $\Phi_t$ is measurable in $(x_0,x_1)$ and differentiable in $t$. The induced law of $x_t$ defines a probability path
\[
\mu_t = (\Phi_t)_\# \pi,
\]
with density $\rho_t$ when it exists.

A standard example is the affine path
\begin{equation}\label{eq:appD_affine_path}
x_t = a(t)x_0 + b(t)x_1,
\end{equation}
where $a(t)$ and $b(t)$ are scalar interpolation functions. In this case, the path derivative is
\begin{equation}\label{eq:appD_path_derivative}
\dot x_t := \frac{d}{dt}x_t = a'(t)x_0 + b'(t)x_1.
\end{equation}
This affine-path view is representative of the broader path-based perspective shared by flow matching and stochastic interpolants \citep{lipman2022flowmatching,albergo2025stochastic}.

\subsection{Conditional versus marginal velocities}
\label{subsec:appD_conditional_vs_marginal}

The object in \eqref{eq:appD_path_derivative} is an endpoint-conditioned velocity: it depends on the sampled pair $(x_0,x_1)$ as well as on time. By contrast, the velocity field appearing in the continuity equation must be a function of the current state and time:
\begin{equation}\label{eq:appD_continuity}
\partial_t \rho_t + \nabla\cdot(\rho_t v_t)=0.
\end{equation}
The corresponding marginally correct velocity field is
\begin{equation}\label{eq:appD_marginal_velocity}
v^*(x,t) = \mathbb{E}[\dot x_t \mid x_t=x,\,t].
\end{equation}

This distinction is essential. Flow matching is implemented using endpoint-conditioned samples and endpoint-conditioned targets, but the learned model
\[
v_\theta(x,t)
\]
is an unconditional function of $(x,t)$. The rest of this appendix explains why these two views are compatible.

\subsection{Conditional flow-matching objective}
\label{subsec:appD_cfm_objective}

The practical conditional flow-matching objective is
\begin{equation}\label{eq:appD_cfm}
\mathcal{L}_{\mathrm{CFM}}(\theta)
=
\mathbb{E}\big[\|v_\theta(x_t,t)-\dot x_t\|^2\big],
\end{equation}
where the expectation is taken over the sampling procedure
\[
t \sim p(t), \qquad (x_0,x_1)\sim \pi, \qquad x_t = \Phi_t(x_0,x_1).
\]
For the affine path \eqref{eq:appD_affine_path}, this becomes
\begin{equation}\label{eq:appD_cfm_affine}
\mathcal{L}_{\mathrm{CFM}}(\theta)
=
\mathbb{E}\Big[
\big\|v_\theta(x_t,t)-\big(a'(t)x_0+b'(t)x_1\big)\big\|^2
\Big].
\end{equation}

This objective is practical because the target $\dot x_t$ is explicitly computable from sampled endpoint pairs.

\subsection{Why the population minimizer is a conditional expectation}
\label{subsec:appD_l2_projection}

We now derive the optimal regression target. Let
\[
Z := (x_t,t), \qquad Y := \dot x_t.
\]
Then the conditional flow-matching loss can be written abstractly as
\[
\mathcal{L}(v)
=
\mathbb{E}\big[\|v(Z)-Y\|^2\big],
\]
where the minimization is over measurable functions of $Z$.

By the standard $L^2$ projection identity,
\begin{align}
\mathbb{E}\big[\|v(Z)-Y\|^2\big]
&=
\mathbb{E}\big[\|v(Z)-\mathbb{E}[Y\mid Z]\|^2\big]
+
\mathbb{E}\big[\|\mathbb{E}[Y\mid Z]-Y\|^2\big].
\label{eq:appD_projection}
\end{align}
The second term is independent of $v$, so the minimizer is achieved by
\begin{equation}\label{eq:appD_l2_minimizer}
v^*(Z)=\mathbb{E}[Y\mid Z].
\end{equation}
Substituting back the definitions of $Z$ and $Y$ yields
\begin{equation}\label{eq:appD_optimal_velocity}
v^*(x,t)=\mathbb{E}[\dot x_t\mid x_t=x,\,t].
\end{equation}
This proves Proposition~\ref{prop:fm_velocity}.

\subsection{Interpretation of the projection formula}
\label{subsec:appD_projection_interpretation}

The decomposition \eqref{eq:appD_projection} shows that conditional flow matching is an ordinary regression problem. The model cannot recover the full endpoint-conditioned derivative $\dot x_t$ because it only observes $(x_t,t)$, not $(x_0,x_1)$ directly. The best it can do in mean-squared error is therefore the conditional expectation of $\dot x_t$ given the information available to it.

This is the basic reason endpoint-conditioned supervision is mathematically compatible with learning an unconditional field $v_\theta(x,t)$ \citep{lipman2022flowmatching}. More recent work has emphasized that this regression viewpoint can be generalized into broader path-based frameworks for fast and consistency-style generative modeling \citep{boffi2025flowmap}.

\subsection{Connection to the continuity equation}
\label{subsec:appD_continuity_connection}

The velocity field \eqref{eq:appD_optimal_velocity} is not merely the best regression target in an abstract statistical sense. It is also the velocity field that is consistent with the induced probability path.

Let $\varphi:\mathbb{R}^d\to\mathbb{R}$ be a smooth test function. Then
\begin{align}
\frac{d}{dt}\mathbb{E}[\varphi(x_t)]
&=
\mathbb{E}\big[\nabla \varphi(x_t)\cdot \dot x_t\big] \\
&=
\mathbb{E}\big[\nabla \varphi(x_t)\cdot \mathbb{E}[\dot x_t\mid x_t,t]\big] \\
&=
\mathbb{E}\big[\nabla \varphi(x_t)\cdot v^*(x_t,t)\big].
\label{eq:appD_weak_continuity}
\end{align}
Equation \eqref{eq:appD_weak_continuity} is the weak form of the continuity equation
\[
\partial_t \rho_t + \nabla\cdot(\rho_t v^*)=0.
\]
Thus the same conditional expectation that arises from the $L^2$ regression argument is also the correct transport velocity for the induced marginal path.

\subsection{Affine paths}
\label{subsec:appD_affine_paths}

For the affine interpolation \eqref{eq:appD_affine_path}, the target derivative is
\[
\dot x_t = a'(t)x_0+b'(t)x_1.
\]
Therefore the marginal optimal velocity is
\begin{equation}\label{eq:appD_affine_optimal}
v^*(x,t)
=
\mathbb{E}[a'(t)x_0+b'(t)x_1 \mid x_t=x,\,t].
\end{equation}
This expression is typically not available in closed form for arbitrary couplings $\pi$, which is precisely why the conditional objective \eqref{eq:appD_cfm_affine} is useful: it avoids having to compute the marginal conditional expectation analytically.

\subsection{Dependence on the coupling}
\label{subsec:appD_coupling_dependence}

A crucial point is that the optimal velocity depends on the coupling $\pi$. Two different couplings may have the same endpoint laws $\mu_0$ and $\mu_1$ but induce different conditional expectations
\[
\mathbb{E}[\dot x_t\mid x_t=x,\,t].
\]
Thus the learned velocity is path-dependent and coupling-dependent. This is one of the major conceptual differences from diffusion models, where the forward SDE more tightly constrains the path.

In this sense, flow matching turns path design into an explicit modeling decision rather than a byproduct of a fixed diffusion.

\subsection{Relation to diffusion, probability-flow ODEs, and neighboring frameworks}
\label{subsec:appD_relation_to_diffusion}

Section~\ref{sec:pfode} showed that a forward diffusion induces a probability-flow ODE with velocity
\[
v_{\mathrm{PF}}(x,t)=f(x,t)-\tfrac12 g(t)^2 s_t(x).
\]
Flow matching instead begins by specifying a path and then learns the corresponding velocity field directly. The two viewpoints coincide when the chosen path matches the diffusion marginals and the conditional expectation in \eqref{eq:appD_optimal_velocity} equals the probability-flow velocity.

More broadly, this path-first perspective overlaps substantially with stochastic interpolants, which explicitly treat flows and diffusions within a common interpolation framework \citep{albergo2025stochastic}. It also overlaps conceptually with Bayesian Flow Networks, which provide another iterative probabilistic route to generation and have recently been related to diffusion-style SDE formulations \citep{graves2023bfn,xue2024unifyingbfn}. Thus flow matching is best understood not as an isolated method, but as one point in a larger family of path-based transport models.

\subsection{Path geometry, rectification, and guidance}
\label{subsec:appD_path_geometry}

The practical success of flow matching depends strongly on the geometry of the chosen path. Rectified flow highlights this explicitly by seeking trajectories that are as straight as possible, thereby reducing stiffness and making coarse ODE discretizations more effective \citep{liu2022rectified}. In the same spirit, recent work on guided flow matching shows that conditioning and guidance can interact substantially with path geometry and with the stability of the learned transport \citep{feng2025guidancefm}. This reinforces one of the central themes of the paper: path choice is not merely a technical implementation detail, but a core modeling decision.

\subsection{Takeaway}
\label{subsec:appD_takeaway}

The main conclusion of this appendix is that the practical flow-matching loss is not merely a heuristic regression objective. Its population minimizer is exactly the marginal velocity field
\[
v^*(x,t)=\mathbb{E}[\dot x_t\mid x_t=x,\,t],
\]
which is both the optimal $L^2$ predictor of the path derivative and the velocity field consistent with the continuity equation for the induced path. This is the formal content of Proposition~\ref{prop:fm_velocity}, and it explains why endpoint-conditioned supervision suffices to learn a state-dependent transport field.

%% file: sections/appendixE_measure_theoretic_background.tex
This appendix clarifies the measure-theoretic language used throughout the survey. The main purpose is not to develop a full abstract framework, but rather to explain why it is useful to distinguish probability \emph{laws} from \emph{densities}, and why many statements in diffusion, score-based modeling, and flow matching are most naturally formulated in terms of probability measures and their pushforwards.

\subsection{Probability laws versus densities}

Let $(\mathcal{X},\mathcal{B})$ be a measurable space, typically $\mathcal{X}=\mathbb{R}^d$ equipped with its Borel $\sigma$-algebra. A probability distribution on $\mathcal{X}$ is formally a probability measure
\[
\mu:\mathcal{B}\to[0,1].
\]
If $\mu$ is absolutely continuous with respect to Lebesgue measure, then there exists a density $\rho$ such that
\[
\mu(dx)=\rho(x)\,dx.
\]
In that case, one often identifies the distribution with its density. However, this identification is only valid when such a density exists.

This distinction matters because many constructions in generative modeling are naturally defined at the level of measures, even when densities may be singular, unavailable in closed form, or only defined after smoothing. For this reason, throughout the paper we use
\[
\mu_t \quad \text{for a probability law}, \qquad \rho_t \quad \text{for its density when it exists}.
\]

\subsection{Why measure language is useful in generative modeling}

There are three recurring reasons to use measure-theoretic language.

\paragraph{(i) Pushforwards are fundamentally measure-valued.}
A generative model often maps a simple random input to a more complex sample. If
\[
z\sim \nu, \qquad x = G_\theta(z),
\]
then the induced distribution of $x$ is the pushforward measure
\[
\mu_\theta = (G_\theta)_\# \nu.
\]
This definition makes sense whether or not $\mu_\theta$ admits a density.

\paragraph{(ii) Pathwise statements concern laws, not only densities.}
When we say that an SDE or ODE induces a family of marginals $(\mu_t)$, this is a statement about the one-time laws of a stochastic or deterministic process. If those laws admit densities, we may write $(\rho_t)$ instead; but the primary object is the probability measure.

\paragraph{(iii) Different dynamics may share marginals but not path measures.}
The probability-flow ODE and the forward SDE may have the same one-time marginals while inducing different distributions over trajectories. This distinction is naturally expressed in terms of measures on path space rather than pointwise densities.

\subsection{Pushforward measures}

Let $T:\mathcal{X}\to\mathcal{Y}$ be measurable, and let $\mu$ be a probability measure on $\mathcal{X}$. The pushforward of $\mu$ by $T$, denoted $T_\#\mu$, is the probability measure on $\mathcal{Y}$ defined by
\begin{equation}\label{eq:appE_pushforward}
(T_\#\mu)(A)=\mu(T^{-1}(A))
\qquad\text{for all measurable } A\subseteq \mathcal{Y}.
\end{equation}
Equivalently, for any bounded measurable test function $\varphi$,
\begin{equation}\label{eq:appE_pushforward_test}
\int_{\mathcal{Y}} \varphi(y)\,(T_\#\mu)(dy)
=
\int_{\mathcal{X}} \varphi(T(x))\,\mu(dx).
\end{equation}

This notation is ubiquitous in transport-based generative modeling. For example:
\begin{itemize}
\item in normalizing flows, the learned map transports a base law to a data law;
\item in deterministic ODE flows, the time-$t$ flow map transports $\mu_0$ to $\mu_t$;
\item in flow matching, a path sampler induces intermediate laws by pushing forward a coupling measure.
\end{itemize}

\subsection{Probability paths}

A probability path is simply a family of probability measures
\[
(\mu_t)_{t\in[0,1]}
\]
connecting an initial law $\mu_0$ and a terminal law $\mu_1$. In diffusion and score-based models, this path is usually defined as the family of one-time laws of a forward SDE. In flow matching, the path is often defined more directly through a coupling and an interpolation rule.

If the path is induced by an interpolation map
\[
x_t = \Phi_t(x_0,x_1),
\]
with coupling $\pi(x_0,x_1)$ between $\mu_0$ and $\mu_1$, then the time-$t$ law is
\begin{equation}\label{eq:appE_path_pushforward}
\mu_t = (\Phi_t)_\# \pi.
\end{equation}
This formula is naturally measure-theoretic: it does not require densities, Jacobians, or smoothness.

\subsection{Path measures}

A one-time marginal path $(\mu_t)$ does not by itself determine a unique law on full trajectories. A \emph{path measure} is a probability measure on a suitable path space, for example
\[
C([0,1],\mathbb{R}^d),
\]
the space of continuous trajectories. An SDE induces a probability measure on this space, and an ODE with random initialization does as well.

This distinction is important because different dynamics may share the same one-time marginals while inducing different path measures. This is precisely the situation for:
\begin{itemize}
\item a forward SDE and its associated probability-flow ODE,
\item different couplings in flow matching,
\item Schr\"odinger bridge formulations that optimize over path-space laws.
\end{itemize}

Thus, when we say that two models are equivalent, we must specify the level of equivalence:
\begin{itemize}
\item equality of endpoint laws,
\item equality of all one-time marginals,
\item or equality of path measures.
\end{itemize}
These are increasingly stronger notions.

\subsection{Weak formulations and continuity equations}

Measure-theoretic language is also useful because transport equations can be written in weak form even when densities are unavailable. Suppose $(\mu_t)_{t\in[0,1]}$ is a probability path and $v_t$ is a velocity field. The continuity equation
\[
\partial_t \rho_t + \nabla\cdot(\rho_t v_t)=0
\]
can be interpreted weakly as
\begin{equation}\label{eq:appE_weak_continuity}
\frac{d}{dt}\int \varphi(x)\,\mu_t(dx)
=
\int \nabla \varphi(x)\cdot v_t(x)\,\mu_t(dx)
\end{equation}
for all smooth compactly supported test functions $\varphi$. This formulation remains meaningful even if $\mu_t$ does not admit a density.

Similarly, the Fokker--Planck equation associated with an SDE can be understood weakly through the infinitesimal generator. This is one reason measure-theoretic formulations are natural in stochastic transport problems.

\subsection{Scores require densities}

Unlike pushforwards or weak continuity equations, the score
\[
s_t(x)=\nabla_x \log \rho_t(x)
\]
requires that the law $\mu_t$ admit a sufficiently smooth positive density $\rho_t$. Thus score-based modeling is inherently a density-based construction, even if the broader transport framework can be expressed at the level of measures.

This is one reason diffusive perturbations are so useful: they regularize the distribution enough that densities become well-defined for positive times in many settings. By contrast, a deterministic transport map may push a measure onto a lower-dimensional set or otherwise create singular structures unless additional regularity is imposed.

\subsection{Marginal equivalence versus path equivalence}

Several equivalence statements in the paper should be interpreted carefully:
\begin{itemize}
\item The probability-flow ODE and the forward SDE are equivalent at the level of one-time marginals, not generally at the level of path measures.
\item Conditional flow matching learns the marginally correct velocity field for a prescribed path, but this depends on the chosen coupling and interpolation rule.
\item Schr\"odinger bridges are naturally formulated as optimization problems over path-space measures, not merely over endpoint densities.
\end{itemize}

In this sense, a measure-theoretic viewpoint helps prevent overinterpretation of formal similarities between methods. Two models may share the same endpoint law or the same marginal path while differing substantially in how probability mass moves over time.

\subsection{Takeaway}

The term \emph{measure-theoretic} in the title of this paper does not mean that the survey develops generative modeling from full axiomatic probability theory. Rather, it means that we systematically distinguish between laws and densities, formulate transport using pushforwards and path measures, and interpret equivalence statements at the correct level of generality. This viewpoint is especially useful for unifying diffusion, score-based models, probability-flow ODEs, and flow matching, because these methods differ not only in their objectives and samplers, but also in the sense in which their underlying transports should be considered equivalent.

%% file: sections/appendixF_schrodinger_bridges.tex
This appendix provides background on Schr\"odinger bridges (SBs) and explains why they are relevant to the unified transport viewpoint developed in this survey. At a high level, Schr\"odinger bridge problems seek the most likely stochastic evolution between prescribed endpoint laws, relative to a reference diffusion. This places them naturally between diffusion-based generative modeling, stochastic control, and entropy-regularized optimal transport.

\subsection{From optimal transport to stochastic bridges}

Classical optimal transport asks for a map or coupling that moves a source law $\mu_0$ to a target law $\mu_1$ at minimal transport cost. In its dynamic formulation, one seeks a probability path $(\mu_t)$ and velocity field $v_t$ minimizing a kinetic-energy functional subject to the continuity equation. In this deterministic setting, transport is described by a path of measures and an associated deterministic flow.

Schr\"odinger bridges modify this picture by introducing stochasticity. Rather than searching over deterministic transports, one searches over probability measures on path space that match prescribed endpoint laws while remaining as close as possible to a chosen \emph{reference path measure}. The resulting problem may be viewed as an entropy-regularized version of dynamic optimal transport.

\subsection{Reference process and path-space KL minimization}

Let $\mathbb{Q}$ denote a reference path measure on trajectory space, for example the law of a diffusion process
\begin{equation}\label{eq:appF_reference_sde}
dX_t = f(X_t,t)\,dt + g(t)\,dW_t.
\end{equation}
The Schr\"odinger bridge problem seeks a new path measure $\mathbb{P}$ solving
\begin{equation}\label{eq:appF_sb_problem}
\min_{\mathbb{P}} \ \mathrm{KL}(\mathbb{P}\,\|\,\mathbb{Q})
\qquad
\text{subject to}
\qquad
X_0 \sim \mu_0,\quad X_1 \sim \mu_1.
\end{equation}
Thus the goal is to find the path-space law that satisfies the endpoint constraints while deviating as little as possible, in relative entropy, from the reference process.

This formulation makes the path-measure viewpoint explicit. Unlike a problem stated only in terms of endpoint densities, \eqref{eq:appF_sb_problem} is an optimization over full stochastic evolutions.

\subsection{Interpretation}

The optimization problem \eqref{eq:appF_sb_problem} admits several complementary interpretations.

\paragraph{(i) Most likely bridge.}
Among all stochastic processes that match the prescribed endpoints, the Schr\"odinger bridge is the one that is most likely relative to the reference diffusion.

\paragraph{(ii) Entropic optimal transport.}
The path-space KL penalty plays the role of an entropy regularizer. As the noise level of the reference process tends to zero, one formally recovers deterministic optimal transport in suitable regimes.

\paragraph{(iii) Controlled diffusion.}
The solution may be interpreted as a controlled version of the reference diffusion, where the control modifies the drift so as to satisfy the endpoint constraints.

These viewpoints help explain why Schr\"odinger bridges connect naturally to both stochastic control and diffusion-based generative modeling.

\subsection{Controlled-diffusion formulation}

Under appropriate regularity assumptions, the Schr\"odinger bridge can be represented as a controlled diffusion
\begin{equation}\label{eq:appF_controlled_sde}
dX_t = \big(f(X_t,t) + u_t(X_t)\big)\,dt + g(t)\,dW_t,
\end{equation}
where the control field $u_t$ is chosen so that the resulting process matches the prescribed endpoint marginals while minimizing the path-space relative entropy. Informally, the bridge is obtained by altering the drift of the reference process as little as possible, measured in an entropy or control-energy sense.

This makes SBs conceptually close to diffusion-based generative models: in both cases, one works with a stochastic process and modifies or reverses its drift to induce a desired distributional evolution.

\subsection{Connection to score-based diffusion modeling}

The connection between Schr\"odinger bridges and score-based diffusion models is most transparent when the reference process is itself a diffusion with tractable forward dynamics. In that case:
\begin{itemize}
\item the reference process provides a forward path measure;
\item the bridge solution modifies the drift to satisfy endpoint constraints;
\item reverse-time and score-based quantities naturally appear in the description of the resulting dynamics.
\end{itemize}

This is one reason SBs are often viewed as a generalization of diffusion-style generative modeling. Standard score-based diffusion models typically fix a forward corruption process and learn reverse-time dynamics from data. Schr\"odinger bridge methods instead treat the endpoint-matching problem itself as a path-space optimization problem relative to a reference diffusion \citep{debortoli2021dsb}.

From the perspective of this survey, diffusion and SB methods share the same broad language:
\begin{enumerate}
\item a reference stochastic process,
\item a path of intermediate laws,
\item drift corrections or controls,
\item and generation by transporting probability mass between endpoints.
\end{enumerate}

\subsection{Why SBs matter for this survey}

Schr\"odinger bridges matter for at least three reasons.

\paragraph{(i) They make the path-measure viewpoint explicit.}
Many statements in diffusion modeling are phrased in terms of one-time marginals. SBs remind us that a stochastic generative model is more fundamentally a law on trajectories, not only a family of endpoint or marginal densities.

\paragraph{(ii) They connect diffusion modeling to entropic transport.}
The bridge formulation clarifies that diffusion-like models can be interpreted not merely as reverse-time simulators, but as approximate solutions to a regularized transport problem on path space.

\paragraph{(iii) They suggest principled conditioning mechanisms.}
Because SBs impose endpoint constraints directly at the level of path measures, they provide a natural conceptual framework for conditional generation and inverse problems, where one wants to transport between constrained distributions rather than merely sample unconditionally.

\subsection{Relation to probability-flow ODEs and flow matching}

Schr\"odinger bridges are not identical to probability-flow ODEs or flow matching, but they sit nearby in the conceptual landscape.

\paragraph{Probability-flow ODEs.}
The probability-flow ODE provides a deterministic dynamics with the same one-time marginals as a diffusion. By contrast, an SB is intrinsically a stochastic path-space object. Thus SBs are closer in spirit to diffusion models than to deterministic ODE transports, even though marginal-equivalence ideas remain relevant.

\paragraph{Flow matching.}
Flow matching begins by selecting a probability path and then learning the corresponding velocity field directly. Schr\"odinger bridges instead define a stochastic optimality problem over path measures relative to a reference process. Both emphasize path design, but they do so in different ways: flow matching treats the path as a modeling choice, while SBs derive a path from an optimization principle.

\subsection{Computational viewpoint}

In practice, exact solution of the Schr\"odinger bridge problem is often intractable, and modern algorithms rely on iterative approximations, alternating updates, or score-based parameterizations of forward and backward quantities. The resulting methods can be seen as blending ideas from:
\begin{itemize}
\item diffusion modeling,
\item iterative proportional fitting or Sinkhorn-like updates on path space,
\item stochastic control,
\item and score estimation.
\end{itemize}
For the purposes of this survey, the key point is not the algorithmic taxonomy but the conceptual role of SBs: they provide a principled path-space formulation that helps organize a broader family of diffusion-inspired transport methods.

\subsection{Limitations of the analogy}

Although Schr\"odinger bridges and diffusion-based generative models are closely related, they should not be conflated. In particular:
\begin{itemize}
\item a standard diffusion model is not automatically solving an SB problem exactly;
\item SB formulations involve explicit endpoint constraints relative to a reference path measure;
\item and the optimization criterion is path-space KL minimization, not simply score matching or likelihood maximization.
\end{itemize}
Thus the relationship is best viewed as one of conceptual overlap and possible algorithmic approximation, rather than literal identity.

\subsection{Takeaway}

Schr\"odinger bridges provide a useful conceptual extension of the transport viewpoint in this survey. They make explicit that stochastic generative modeling can be formulated as optimization over path measures relative to a reference diffusion, subject to endpoint constraints. This strengthens the connections between diffusion models, stochastic control, and entropic optimal transport, and helps explain why path design, conditioning, and drift correction play such central roles across modern generative modeling methods \citep{debortoli2021dsb}.